\pdfoutput=1

\documentclass[11pt]{article}

\PassOptionsToPackage{table}{xcolor}
\usepackage{acl} 

\usepackage{times}
\usepackage[percent]{overpic}
\usepackage{latexsym}

\usepackage[T1]{fontenc}

\usepackage[utf8]{inputenc}

\usepackage{microtype}

\usepackage{tikz}
\usetikzlibrary{trees}
\usetikzlibrary{shapes, arrows.meta, positioning, backgrounds, fit}

\usepackage{colortbl}

\usepackage{xcolor}
\definecolor{thedarkblue}{RGB}{0,0,120}
\definecolor{mydarkblue}{rgb}{0,0.08,0.45}
\definecolor{darkblue}{rgb}{0,0.08,180}
\colorlet{TufteRed}{red!80!black}
\definecolor{theblue}{RGB}{0,0,180}
\colorlet{thered}{TufteRed}

\usepackage{microtype}
\usepackage{balance}
\usepackage{booktabs}
\usepackage{tabularx}
\usepackage[most]{tcolorbox}
\usepackage{fontawesome5} 
\usepackage{caption}      

\newtcolorbox{promptbox}[2][]{
  enhanced,
  colback=white!98!blue!2, 
  colframe=blue!70!black,  
  coltitle=white,          
  fonttitle=\bfseries\sffamily,
  title={\faTasks[regular]~\hspace{1mm}\thetcolorbox #2},
  left=2mm, right=2mm, top=1mm, bottom=1mm,
  arc=3mm,                  
  attach boxed title to top left={xshift=3mm, yshift*=-2mm},
  boxed title style={
    colback=blue!80!black,
    size=small,
    sharp corners=south,
    bottom=0.5mm, top=0.5mm,
    left=1mm, right=1mm,
    fontupper=\bfseries\sffamily,
    boxrule=0pt,
    drop shadow
  },
  boxrule=0.9pt,
  drop shadow southeast,
}

\usepackage{amsmath,amssymb,amsthm}

\newcommand{\eat}[1]{\ignorespaces}
\usepackage{comment}

\usepackage{verbatim}

\usepackage{ragged2e}
\usepackage{multirow}
\usepackage{microtype}
\usepackage{balance}
\usepackage{setspace}

\graphicspath{{./}{./graphics/}}
\newcolumntype{H}{>{\setbox0=\hbox\bgroup}c<{\egroup}@{}}
\newcolumntype{R}[1]{>{\RaggedLeft\arraybackslash}}
\newcolumntype{L}[1]{>{\RaggedRight\arraybackslash}}

\AtBeginEnvironment{pmatrix}{\setlength{\arraycolsep}{2pt}}

\DeclareMathOperator{\hugeE}{\mbox{\huge\raise-0.3ex\hbox{E}}}
\DeclareMathOperator{\p}{\mathbb{P}}
\DeclareMathOperator{\hugep}{\mbox{\huge\raise-0.3ex\hbox{$\p$}}}

\usepackage{subfigure}
\usepackage{float}
\usepackage{graphbox}
\usepackage{svg}
\usepackage{nicefrac}
\usepackage{bold-extra}
\newcolumntype{P}[1]{>{\centering\arraybackslash}p{#1}}
\newcolumntype{M}[1]{>{\centering\arraybackslash}m{#1}}

\definecolor{orange}{rgb}{1,0.5,0}
\definecolor{graynode}{RGB}{20,20,20}
\definecolor{crimsonred}{RGB}{220,20,60}
\definecolor{darkgraynode}{gray}{0.5}
\definecolor{lightgraynode}{gray}{0.8}

\usepackage{rotate}
\usepackage{adjustbox}
\usepackage{array}
\usepackage{capt-of}
\usepackage{tabulary}
\usepackage{setspace}
\usepackage{amssymb}
\usepackage{mathtools}
\usepackage{pifont}
\usepackage{tabularx}

\newcommand{\cmark}{\ding{51}}
\newcommand{\xmark}{\ding{55}}

\definecolor{gray}{RGB}{0.7,0.7,0.7}
\definecolor{greencm}{RGB}{0,153,0}

\newcommand{\cm}{ {\color{greencm}\normalsize\cmark}}

\newcommand{\xm}{ {\color{red}\normalsize\xmark}}

\definecolor{plotblue}{RGB}	{30,144,255}
\definecolor{plotgreen}{RGB}	{50,205,50}
\definecolor{plotred}{RGB}	{220,20,60}

\definecolor{myyellow}{RGB}{255,255,204}
\definecolor{myred}{RGB}{255,204,204}
\definecolor{myblue}{RGB}{0,200,255}
\definecolor{mygreen}{RGB}{80,220,80}
\definecolor{Green}{RGB}{50,205,50}
\usepackage{color}

\newcommand*\hrulefillvar[1][0.4pt]{\leavevmode\leaders\hrule height#1\hfill\kern0pt}

\usepackage{subfigure}
\usepackage{nicefrac}

\DeclareMathAlphabet{\mathbcal}{OMS}{cmsy}{b}{n}
\usepackage{amsmath}
\usepackage{mathrsfs}
\usepackage{comment}

\usepackage{bm}
\usepackage{bbm}
\usepackage{amssymb}
\usepackage{amsthm}

\theoremstyle{definition}
\newtheorem{definition}{Definition}[section]

\usepackage{paralist}
\newcolumntype{C}{ >{\centering\arraybackslash} m{4cm} }
\providecommand{\rotateDeg}{90}
\definecolor{verylightgreennew}{RGB}	{220,255,220}
\definecolor{verylightrednew}{RGB}		{255, 230, 230}
\definecolor{verylightreddarker}{HTML} {FFCBCB} 
\definecolor{verylightrednewlighter}{RGB}		{255, 229, 239}
\definecolor{lightgraynew}{rgb}{0.95,0.95,0.95}
\definecolor{newgray}{RGB}{0.3,0.3,0.3}
 
\providecommand{\cellszlg}{0.39cm} 
\providecommand{\cellszsm}{0.40cm} 
\renewcommand{\xm}{{\color{verylightreddarker}\normalsize\xmark}}
\newcommand\BBBBB{\rule[1.6ex]{0pt}{1.6ex}}
\newcommand\BBBnew{\rule[-2.5ex]{0pt}{0pt}} 
\newcommand\BBBBBB{\rule[-1.1ex]{0pt}{0pt}} 
\newcommand{\sysName}[1]{{\sf
\BBBBBB
#1
}}

\providecommand{\cellno}{
\BBBBB
\xm
\cellcolor{verylightrednew}}
\providecommand{\cellyes}{
\BBBBB
\cm
\cellcolor{verylightgreennew}
}

\definecolor{thedarkblue}{RGB}{0,0,120} 
\definecolor{mydarkblue}{rgb}{0,0.08,0.45} 

\usepackage{hyperref}
\hypersetup{%
    colorlinks=true,
    linkcolor=mydarkblue,
    citecolor=mydarkblue,
    filecolor=mydarkblue,
    urlcolor=mydarkblue}

\usepackage{makecell}

\definecolor{googleblue}{HTML}{4285F4}
\definecolor{googlered}{HTML}{DB4437}
\definecolor{googlepurple}{HTML}{A142F4} 
\definecolor{googlegreen}{HTML}{0F9D58}
\definecolor{googleyellow}{HTML}{F4B400} 
\definecolor{googleorange}{HTML}{FBBC05} 
\definecolor{googlecyan}{HTML}{34A853} 
\definecolor{googlegray}{HTML}{9AA0A6} 
\definecolor{googlepink}{HTML}{EA4335} 
\definecolor{googlelightblue}{HTML}{7BAAF7} 

\title{A Survey on LLM-based Conversational User Simulation}

\author{Bo Ni$^{1}$, 
Leyao Wang$^{3}$,
Yu Wang$^{4}$,
Branislav Kveton$^{2}$,
Franck Dernoncourt$^{2}$,
Yu Xia$^{5}$,\\
\textbf{
Hongjie Chen$^{6}$,
Reuben Leura$^{7}$,
Samyadeep Basu$^{2}$,
Subhojyoti Mukherjee$^{2}$,}\\
\textbf{
Puneet Mathur$^{2}$,
Nesreen Ahmed$^{8}$,
Junda Wu$^{5}$,
Li Li$^{9}$,
Huixin Zhang$^{10}$,
Ruiyi Zhang$^{2}$,}\\
\textbf{
Tong Yu$^{2}$,
Sungchul Kim$^{2}$,
Jiuxiang Gu$^{2}$,
Zhengzhong Tu$^{10}$,
Alexa Siu$^{2}$,
Zichao Wang$^{2}$,
}\\
\textbf{
David Seunghyun Yoon$^{2}$,
Nedim Lipka$^{2}$,
Namyong Park,
Zihao Lin$^{11}$,
Trung Bui$^{2}$,}\\
\textbf{
Yue Zhao$^{9}$,
Tyler Derr$^{1}$,
Ryan A. Rossi$^{2}$}\\ \\
$^{1}$Vanderbilt University, 
$^{2}$Adobe Research, 
$^{3}$Yale University, 
$^{4}$University of Oregon, \\
$^{5}$University of California San Diego,
$^{6}$Dolby Laboratories, 
$^{7}$University of California, Berkeley, \\
$^{8}$Cisco AI Research, 
$^{9}$University of Southern California, 
$^{10}$Texas A\&M University, 
$^{11}$UC Davis
}

\begin{document}
\maketitle

\begin{abstract}
User simulation has long played a vital role in computer science due to its potential to support a wide range of applications. Language, as the primary medium of human communication, forms the foundation of social interaction and behavior. Consequently, simulating conversational behavior has become a key area of study. Recent advancements in large language models (LLMs) have significantly catalyzed progress in this domain by enabling high-fidelity generation of synthetic user conversation. In this paper, we survey recent advancements in LLM-based conversational user simulation. We introduce a novel taxonomy covering user granularity and simulation objectives. Additionally, we systematically analyze core techniques and evaluation methodologies. We aim to keep the research community informed of the latest advancements in conversational user simulation and to further facilitate future research by identifying open challenges and organizing existing work under a unified framework.

\end{abstract}

\begin{figure}[t!]
\centering
\vspace{2mm}
\includegraphics[width=0.97\linewidth]{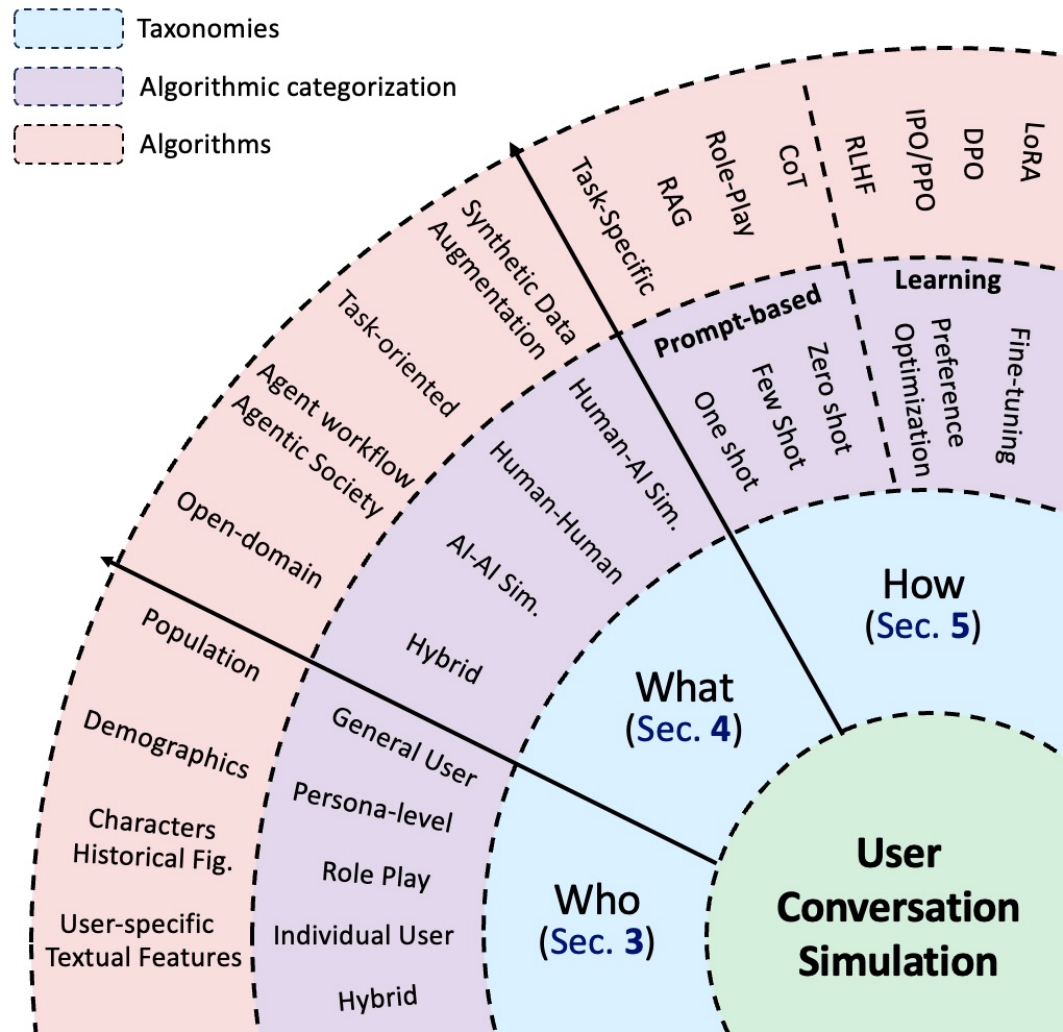}
\caption{
Overview of the proposed taxonomies for user conversation simulation.
}
\vspace{-0.1in}
\label{fig:overall-taxonomy}
\end{figure}

\section{Introduction}
\label{sec:introduction}

User simulation has been an active area of research for decades.\footnote{Simulation is often used interchangeably with the term generation, creation, and so on. Similarly, the term conversation is sometimes used interchangeably with dialogue, chat, multi-turn interaction, among others.} From early simulation games such as \textit{The Sims} \cite{sims} to agent-based environments powered by large language models \cite{park2023generative}, the goal has consistently been to create realistic user proxies that support diverse applications. The foundations of user simulation trace back to classic models of user preferences, such as the Bradley-Terry-Luce \cite{bradley1952rank} and Plackett-Luce \cite{plackett1975analysis} models. In these models, simulation occurs implicitly through statistical models trained to emulate user preferences and behaviors from observational data, and they remain influential until today in reinforcement learning from human feedback \cite{ouyang2022training} and direct preference optimization \cite{rafailov2023direct}. Other statistical models of user behavior, such as collaborative filtering \cite{schafer2007collaborative}, low-rank matrix factorization \cite{koren2009matrix}, Bayesian ranking \cite{rendle2012bpr}, and click models \cite{chapelle2009dynamic} have been widely adopted to capture user behavior and facilitate personalized experiences.

The emergence of large language models (LLMs) has significantly transformed the landscape of user simulation in two key ways: first, it enables general-purpose simulation across a wide range of tasks and domains; and second, it drastically lowers the barrier to generating high-quality contextually rich simulated interactions via prompt engineering of pretrained models. In contrast, past frameworks for user simulation in recommender systems \cite{rohde2018recogym,ie2019recsim} required large amounts of user data to train user simulators and were tailored only to a particular form of recommendations. Unlike simulation in more deterministic domains such as physics~\cite{chan1999simulation}, climate~\cite{change2021climate}, or epidemiology~\cite{ferguson2006strategies}, user simulation must grapple with the inherent complexity, variability, and nuance of human behavior. LLMs offer a powerful tool for capturing this complexity—generating coherent, adaptive, and goal-directed behavior without task-specific supervision, making them well-suited for simulating user interactions at scale.

A growing body of work has leveraged LLMs to simulate different user behaviors. In retrieval-augmented generation (RAG)~\cite{lewis2020retrieval, salemi2023lamp}, the LLM generates a summary of retrieved documents by a classic retrieval system, and thus simulates how a real user would review retrieved knowledge and internalize it. Chain-of-thought (CoT) reasoning~\cite{wei2022chain} mimics users' internal reasoning processes, producing step-by-step interactions that reflect human-like deliberation. The LLM-as-a-judge \cite{gu2024survey} simulates how users evaluate responses in terms of helpfulness or correctness, for instance. More domain-specific simulation frameworks have emerged to improve downstream applications such as search, recommendation, and task-oriented dialogue, enabling systems to adapt to user intent and context with greater fidelity. For instance, BASES~\cite{ren2024bases} generates diverse user profiles and simulates large-scale web search behavior, addressing data scarcity and privacy concerns. USimAgent~\cite{zhang2024usimagent} further replicates user querying, clicking, and session behaviors with strong alignment to real-world interactions. To ensure fidelity, Breuer et al.~\cite{breuer2024validating} propose methods for validating synthetic usage data in data-sparse environments. Finally, Balog and Zhai~\cite{balog2025user} present an integrated framework that advances both user modeling and system evaluation through generative AI.

While these methods provide powerful mechanisms for simulating user behaviors, realistic user interactions often require explicitly modeling the interactive, conversational nature of communication. Language, as the primary medium of human interaction, plays a central role in shaping human behavior~\cite{fitch2010social, tomasello2010origins}. LLMs, with their unprecedented language capability, therefore offers a transformative opportunity for conversational user simulations. For example, ~\citet{zhang2025llm} showed that LLM-generated conversational feedback, such as synthetic user comments, can directly improve recommender system performance. Similarly,~\citet{sekulic2024reliable} demonstrated that conversations can better align systems with user needs and improve satisfaction in task completion. Despite these promising results and a growing body of work in language-based user simulation~\cite{hazrati2022simulating, wang2023user, zhang2025llm}, a dedicated survey that systematically organizes and analyzes the sub-field of conversational user simulation is absent. 

In this survey, we aim to fill this gap by providing a comprehensive overview and unified taxonomy of conversational user simulation. As outlined in Table~\ref{tab:taxonomy-techniques} in the Appendix, we first define the key components and scope of this emerging area. We then organize the survey to answer three fundamental questions: \textit{(1) \textbf{Who} is being simulated?} (\S\ref{sec:who}), \textit{(2) \textbf{What} is being simulated?} (\S\ref{sec:what}), and \textit{(3) \textbf{How} is the conversation simulated?} (\S\ref{sec:how}). With this organized discussion, we aim to highlight key research trends and pinpoint open challenges, fostering future research in this area. For more details on related surveys and how our work differs, please refer to Appendix~\ref{app:survey_diff}.

\section{Problem Definition}
We formally define a conversation as a sequence of turns between two or more participants. Let $\mathcal{P} = \{p^1, p^2, \dots, p^N\}$ be the set of $N$ participants in a conversation. Note that these participants can be of different types, including human users ($\mathcal{P}_U$), systems ($\mathcal{P}_S$), or other agent types, allowing for user-user, user-system, and multi-party conversation simulations.

A conversation, $C$, is a temporally ordered sequence of $T$ turns: $C = (c_1, c_2, ..., c_{T})$ where each turn $c_t$ for $t \in [1, T]$ is a tuple containing the speaker and their utterance:
\[
c_t = (p_t^i, u_t)
\]
where $p_t^i \in \mathcal{P}$ denotes the participant  $p^i$ speaking at turn $t$, and $u_t$ is the utterance they produce from a vocabulary $\mathcal{V}$.

The core of conversational simulation is to model the behavior of one or more target participants. Let $C_{t-1}$ be the conversational history and $C_{t-1} = (c_1, \dots, c_{t-1})$. Additionally, let $\Psi_{p^i}$ be the context of participant $p^i$ that can include their demographics, domain knowledge, or personal context, etc. Then, the fundamental task of a conversational simulator is to generate a participant's next utterance by modeling the probability distribution:
\[
P(u_t | C_{t-1}, \Psi_{p^i}) 
\] 

\section{Who: Simulated Users and Interactions}
\label{sec:who}

\begin{figure}
    \centering
    \includegraphics[width=0.98\linewidth]{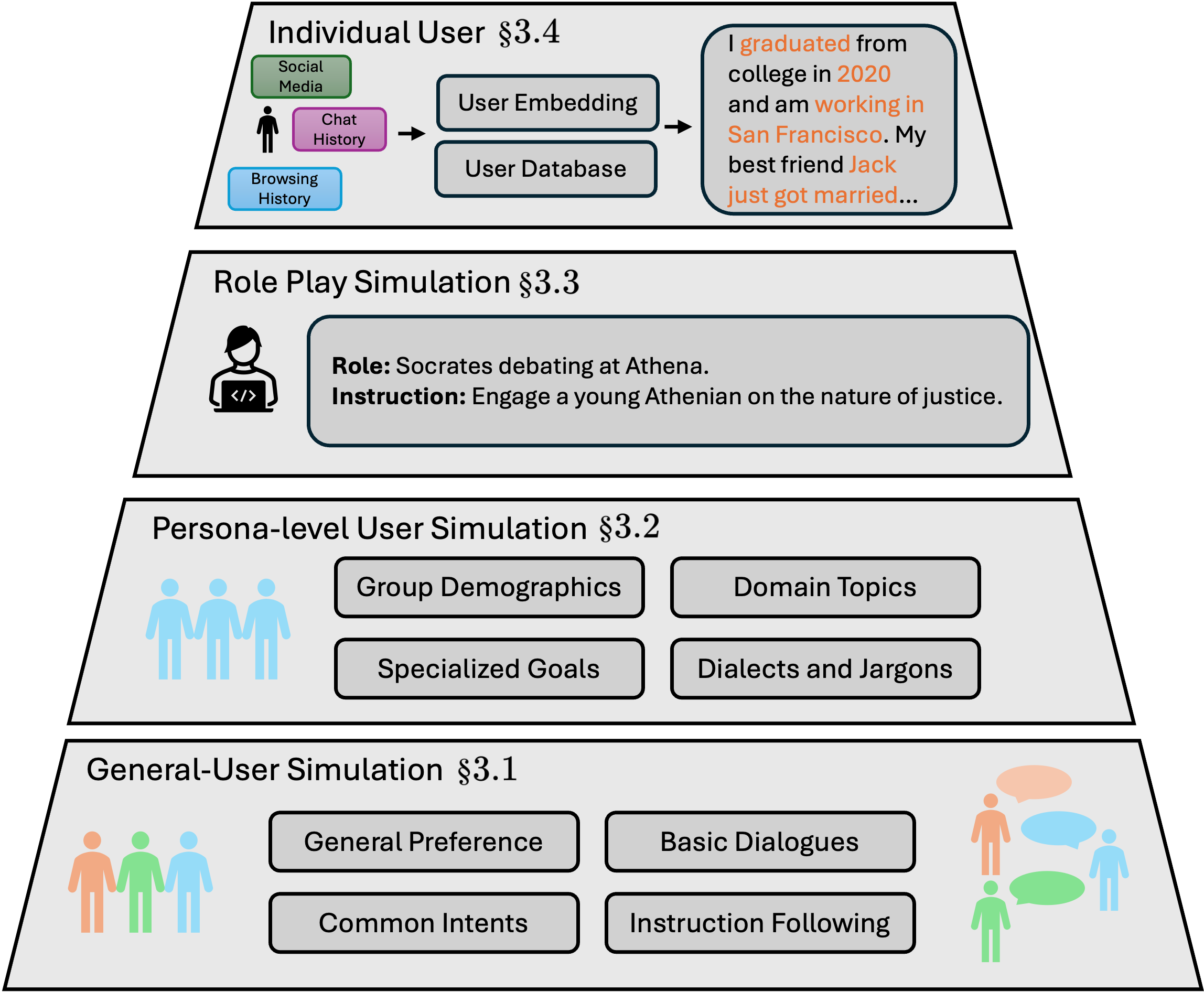}
    \vspace{-0.05in}
    \caption{Overview of the proposed taxonomy for \textbf{who} is being simulated, i.e., the target of simulation.}
    \label{fig:who_fig_general}
    \vspace{-0.15in}
\end{figure}

The first core question that we explore is the target of simulation. As illustrated in Figure \ref{fig:who_fig_general}, this section discusses simulation of conversations going from general users (Sec.~\ref{sec:who-general-users}), to persona-level simulation of conversations (Sec.~\ref{sec:who-persona-based-sim}), to more role-playing simulation (Sec.~\ref{sec:who-role-based-sim}), to the most fine-grained individual user conversation simulation 
(Sec.~\ref{sec:who-user-level-sim}). In each section, we will first formally define the corresponding simulation target, and then introduce relevant works in the area.

\subsection{General User Simulation} \label{sec:who-general-users}
General user simulation models conversations from a broad population perspective, as seen in models like ChatGPT~\cite{achiam2023gpt}. 

\begin{definition}[General User Simulation]
\label{def:general_user}
The persona $\Psi_p^{default}$ is considered \emph{default}, representing an average user sampled randomly from the general population $\mathcal{P}_U$. No distinguishing characteristics are provided in the prompt.
\end{definition}

Recent advances improve interaction quality through multi-turn optimization~\cite{xiong2024building}, exploration-based learning~\cite{song2024trial}, and diverse user simulation~\cite{dhole2024kaucus}. For a detailed discussion of trajectory-level fine-tuning, failure-aware exploration, segment-based optimization, and reinforcement-based training approaches, please refer to Appendix~\ref{append:general-user}.

\subsection{Persona-level User Simulation} \label{sec:who-persona-based-sim}

Persona-level user simulation models users at the demographics level~\cite{chen2024persona}, thereby going beyond general user simulations.

\begin{definition}[Persona-level User Simulation]
The persona $\Psi_p$ is explicitly defined by a set of $m$ attributes: $\Psi_p = \{\psi_1, \psi_2, \dots, \psi_m\}$, where each $\psi_i$ corresponds to a demographic, interest, or stylistic feature.
\end{definition}

Recent work explores a range of techniques for persona grounding, including demographic prompting~\cite{hu2024quantifying}, psychometric modeling~\cite{wang2025user, ji2024persona}, trait-infused architectures~\cite{huang2024orca, yang2025psyplay}, and activation-level control~\cite{zhu2024personality}. These methods have shown promise but also raise concerns around fairness and bias~\cite{li2025llm, deshpande2023toxicity}. This underscores the need for trustworthy simulations. For a more detailed discussion of these approaches and their implications, please refer to Appendix~\ref{append:personal-simulation}.

\subsection{Role Play Simulation} \label{sec:who-role-based-sim}
Unlike persona-level simulation (\S\ref{sec:who-persona-based-sim}), which relies on fixed traits, role play simulation offers greater flexibility by modeling real or fictional individuals. For instance, while persona-level simulation may represent a generic ``20-year-old programmer,'' role play simulation can emulate ``Mark Zuckerberg at 20,'' including his unique style and historical context.
\begin{definition}[Role Play Simulation]
Let $\Phi_\theta$ denote an LLM with parameters $\theta$ and let $\mathcal{M}$ be its latent manifold. 
Given an identity handle $h$ (e.g., ``Mark Zuckerberg at 20'') expressed in natural language, the model induces an implicit embedding
\[
I \;\coloneqq\; E_\theta(h) \in \mathcal{M},
\]
where $E_\theta:\Sigma^\ast \!\to\! \mathcal{M}$ is the encoder realized by $\Phi_\theta$. 
A role-play persona is the conditional output distribution
\[
\Psi_p \;\coloneqq\; p_\theta(y \mid x, I) \;=\; p_\theta\!\big(y \mid x, E_\theta(h)\big),
\]
i.e., the model’s behavior when inputs $x$ are conditioned on the latent identity $I$. 
\end{definition}

This paradigm leverages LLMs' implicit knowledge to simulate rich character traits~\cite{serapio2023personality, wang2023rolellm}, enhanced by prompting~\cite{wang2025user}, fine-tuning~\cite{shao2023character}, and self-play~\cite{ji2025enhancing}. Applications span storytelling~\cite{wu2024role}, social simulation~\cite{wang2025user}, and agent memory modeling~\cite{fan2025if}. For a full discussion, see Appendix~\ref{append:role-play-simulation}.

\subsection{Individual User Simulation} \label{sec:who-user-level-sim}

Individual user simulation represents the most fine-grained level of conversational modeling in our taxonomy. Unlike role-play simulations, which rely on \textit{implicit} predefined characters or figures, individual user simulations are grounded in the \textit{explicit}, often dynamic, personal context of users.

\begin{definition}[Individual User Simulation]
The persona $\Psi_p$ is derived from the full personal history $\mathcal{H}_p$ of participant $p$, including chat logs, documents, and interaction histories, i.e., $\Psi_p = \mathcal{H}_p$.
\end{definition}

Techniques range from profile injection~\cite{zhang2018personalizing, jang2022call} and dialogue history modeling~\cite{li2021dialogue} to multi-session memory~\cite{xu2021beyond, chhikara2025mem0} and real-world trait grounding~\cite{yamashita2023realpersonachat, gao2023livechat}. For a detailed discussion of the related work, please refer to Appendix~\ref{append:individual-simulation}.

\subsection{Hybrid User Simulation}  
\label{sec:who-hybrid}

While general user, persona-level, role play, and individual user simulations are conceptually distinct, they often overlap in practice. Large-scale models such as GPT~\cite{achiam2023gpt} and LLaMA~\cite{touvron2023llama} naturally blend these paradigms, often exhibiting emergent persona and role-play behaviors~\cite{wang2023rolellm}. Recent works have demonstrated the use of GPT-4 and related models to simulate dialogues for evaluation, data augmentation, and conversational benchmarks~\cite{park2023generative, guo2024large}. However, these studies typically adopt a single simulation target (e.g., role play or individual user emulation) without explicitly formalizing hybrid combinations. Thus, the systematic treatment of hybrid user simulation remains limited. The hybrid user simulation can be particularly relevant for multi-agent interactions, as emergent behavior depends on balancing generic user roles with individual variation. 
\section{What: Simulation Objectives}  
\label{sec:what}  

This section defines the \emph{objectives} of LLM-based user simulation by focusing on \emph{interaction patterns}, rather than applications (\S\ref{sec:apps}). We categorize these into four paradigms: \emph{Human–AI}, \emph{Human–Human}, \emph{AI–AI}, and \emph{Many-Human–AI} simulation. Let an interaction trajectory $\tau$ be defined as 
\begin{equation}
    \tau = ((u_1, v_1), \dots, (u_T, v_T))
\end{equation}
where $T$ is the number of turns in the interaction. The paradigms in the rest of this section will be defined following the definition.

\subsection{Human–AI Simulation}
\label{sec:what-human-ai}

Human–AI simulation models turn-based interactions where a human prompts and the AI responds, aiming not just to mimic behavior but to create realistic contexts for evaluating AI capabilities. We define a \emph{Human-AI simulation} as any trajectory where $u, v$ is defined as human utterance and AI response, respectively. At each turn \(t\), the human utterance \(u_t\) is generated according to one of the ``Who'' levels:
    \[
      h_t \sim P\bigl(C_{t-1}, \,\Psi_{p},\bigr),
    \]
    where \(\Psi_{p}\) is the corresponding profile defined in the previous sections. The AI response \(v_t\) is generated by a fixed model with $\Psi_p^{default}$ being the default profile.

Recent approaches generate synthetic human–AI dialogues to reduce reliance on expensive annotation, including Self-Instruct~\cite{wang2022self}, WizardLM~\cite{xu2024wizardlm}, and Auto Evol-Instruct~\cite{zeng2024autoevol}. Other methods target broad coverage~\cite{li2024synthetic, zhang2024multimodal}, domain adaptation~\cite{das2024synthetic, rachidi2025design}, or task specificity~\cite{wang2024codeclm, patel2024datadreamer}. For a detailed discussion, see Appendix~\ref{append:human-ai}.

\subsection{Human–Human Simulation}
\label{sec:what-human-human}

Human–Human simulation models conversations between two human participants, aiming to replicate natural dialogue patterns~\cite{yamashita2023realpersonachat, dinan2018wizard}. This paradigm supports the development of agents that can maintain consistent personas and engage in grounded interactions. We define a \emph{Human–Human simulation} as any trajectory where each participant's utterances are initialized by human input at some ``Who'' level:
\[
\begin{aligned}
  u_t &\sim P\bigl(C_{t-1}, \,\Psi_{p_u},\bigr),\\
  v_t &\sim P\bigl(C_{t-1}, \,\Psi_{p_v},\bigr),
\end{aligned}
\]
Both sides use their own profiles \(\Psi_{p_*}\) to simulate realistic two‑party human dialogue.

Key datasets include PersonaChat~\cite{zhang2018personalizing}, Wizard-of-Wikipedia~\cite{dinan2018wizard}, EmpatheticDialogues~\cite{rashkin2018towards}, and MultiWOZ~\cite{budzianowski2018multiwoz}. Scalable alternatives such as self-play bootstrapping~\cite{shah2018bootstrapping} reduce reliance on manual curation. For an extended discussion, see Appendix~\ref{append:human-human}.

\subsection{AI–AI Simulation}
\label{sec:what-ai-ai}

AI–AI simulation models conversations where both participants are autonomous AI agents, interacting without human input~\cite{park2023generative, ren2024emergence}. This paradigm enables scalable data generation and the study of emergent behaviors. We define \emph{AI-AI} simulation as any trajectory where two AI 
agents \(A_1,A_2\) converse \emph{without} ongoing human input.  The only human contribution is a \emph{general seed prompt} \(\mathcal{Q}\), and thereafter
\[
a^1_t,\,a^2_t \;\sim\; P\bigl(\mathcal{Q},\,a^1_{1:t-1},\,a^2_{1:t-1}\bigr),
\]
alternating turns.  No persona, role, or individual‐level profiles are specified at the entity level. Although certain level of simulation (persona, role-play, individual) might be assigned based on the seed prompt $\mathcal{Q}$.

Recent work explores emergent social behaviors~\cite{park2023generative, ren2024emergence}, collaborative task-solving~\cite{li2023camel, wu2023autogen}, and adversarial debate~\cite{du2023improving, rennard2025bias, hua2024game}. For a comprehensive discussion, see Appendix~\ref{append:AI-AI}.

\subsection{Many-Human–AI Simulation}
\label{sec:what-many-human-ai}

Many-human–AI simulation generalizes human–AI interaction to multi-user settings, where multiple humans engage with one or more AI agents toward a shared objective. This paradigm captures both individual behavior and group dynamics in collaborative environments.

\begin{definition}[Many‑Human–AI Simulation]
Let $\mathcal{U} = \{u^1,\dots,u^n\}$
be a set of $n$ human participants, each endowed with a profile
\(\Psi_{u^i}\),
and let $A$ be an AI system that follows general user behavior following definition~\ref{def:general_user}.
A \emph{Many‑Human–AI simulation} is any dialogue trajectory
\[
\tau = \{\tau^1, \tau^2, \dots, \tau^n\}
\]
where
\[
\tau^j = \bigl\{(a_t,\,u^j_t)\bigr\}_{t=1}^T,
\]
Each turn \(t\) is produced as follows:
\[
u^j_t \sim P\bigl(C_{t-1},\,\Psi_{u^j}\bigr)\]
\[
a_t \sim P\bigl(u^{1:t},\,\mathcal{Q}\bigr)
\]
where $\mathcal{Q}$ is the default prompt to align the AI to general user profile.
\end{definition}

Recent work explores collaborative roles~\cite{klieger2024chatcollab}, proxy participation~\cite{leong2024dittos}, and group dialogue simulation~\cite{mao2024multi}. Despite the advancement in AI-AI simulations~\cite{li2023camel, wu2023autogen} and Human-AI~\cite{wang2022self} simulations, general-purpose frameworks for simulating many-human–AI settings remain limited. For an extended discussion, see Appendix~\ref{append:many-human}.

\subsection{Hybrid Simulation}
\label{sec:what-hybrid}

Hybrid simulation blends multiple paradigms (i.e., Human–Human, AI–AI, and Human–AI) within the same environment, reflecting the complexity of real-world interactions. For example, sandbox environments like Smallville~\cite{park2023generative} simulate AI–AI communities, yet individual dialogues within them resemble Human–Human exchanges. 
Despite its prevalence, hybrid simulation remains under-theorized. We advocate for systematic frameworks to model and benchmark mixed-interaction settings, essential for socially adaptable AI.

\section{How: Techniques and Methodologies}
\label{sec:how}

To generate synthetic user conversations, it is crucial to incorporate diverse types of contextual knowledge, each requiring distinct representations. For example, static persona facts guide tone and content~\cite{li2016persona}, while long-term preferences use memory-based retrieval~\cite{madotto2018mem2seq}. We discuss integration strategies for these heterogeneous information in the next section.

\subsection{Prompt‑Based Simulation}\label{sec:how-prompt}

Prompt-based simulation formulates user generation as conditional language modeling:
\[
  u_t \sim P(C_{t-1}, \Psi_p, \mathcal{P}),
\]
where $\mathcal{P}$ is the optional in‑context
examples. Existing work falls into two complementary
tracks:
method-driven(zero-shot/few-shot and chain-of-thought) and content-driven(persona/role-play and task-specific prompts). Zero-/few-shot prompts enable scalable simulation with minimal examples~\cite{terragni2023context, zhang2024generative, wang2025user}, while Chain-of-Thought improves coherence via step-by-step reasoning~\cite{luo2024duetsim}. Persona and task-specific prompts guide tone and domain behavior~\cite{shanahan2023role, abbasiantaeb2024let, kong2023platolm, kiesel2024simulating}. See \S\ref{append:how-prompt} for more details.

\subsection{RAG} \label{sec:how-rag}
Retrieval-augmented generation (RAG) enhances user simulation by conditioning responses on external knowledge:
\[
u_t \sim P(C_{t-1}, \Psi_p, \mathcal{R}(C_{t-1}, \Psi_p)) 
\]
where $\mathcal{R}(\cdot)$ retrieves context to enhance realism, relevance, and personalization of behaviors.

RAG methods vary by retrieval trigger mechanism. Always-on approaches~\cite{shimadzu2025retrieval}, retrieve at every turn prepend relevant passages. Adaptive methods~\cite{wang2024adaptive} use a learned classifier to decide when retrieval is necessary to improve efficiency. Goal/State-driven approaches~\cite{zhu2025llm} retrieve based on internal user memory for more personalized generation. For further examples, see Appendix~\ref{append:how-rag}.

\subsection{Fine-tuning}
\label{sec:how-fine-tuning}

Fine-tuning adapts a pretrained language model into a user simulator by updating its parameters $\boldsymbol{\Theta}$ on a dataset of user dialogues $\mathcal{D}$.  
Formally, we define $\mathcal{D} = \{(C_{t-1}, \Psi_p, \mathcal{I}, u_t)\}_{t=1}^T$, where $C_{t-1}$ denotes the conversation history up to turn $t-1$, $\Psi_p$ is the persona description associated with the user, $\mathcal{I}$ represents optional task-specific instructions, and $u_t$ is the ground-truth user utterance at turn $t$.  

During training, the fine-tuned model is optimized to generate user utterances conditioned on the given context:
\[
u_t \sim P_{\boldsymbol{\Theta}'}(C_{t-1}, \Psi_p, \mathcal{I}),
\]
where the updated parameters $\boldsymbol{\Theta}'$ are obtained by maximizing a supervised fine-tuning objective:
\[
\boldsymbol{\Theta}' = \arg\max_{\boldsymbol{\Theta}} \mathcal{L}_{\mathrm{FT}}(\boldsymbol{\Theta}; \mathcal{D}).
\]
Here, $\mathcal{L}_{\mathrm{FT}}$ typically corresponds to the cross-entropy loss between the predicted and ground-truth user responses $u_t$.

We group fine-tuning strategies into three types. Full-model supervised methods retrain all parameters on in-domain data, as in DAUS~\cite{sekulic2024reliable}, SoulChat~\cite{chen2023soulchat}, and MuPaS~\cite{wang2024multi}. Parameter-efficient approaches like ESC-Role~\cite{zhao2024esc}, BiPO~\cite{cao2024personalized}, and SRA-guided LoRA~\cite{madani2024steering} use adapters or activation steering to preserve efficiency. Interactive methods optimize simulators via interaction feedback: UGRO~\cite{hu2023unlocking} applies reward modeling with PPO, while PlatoLM~\cite{kong2023platolm} leverages large-scale self-play before assistant fine-tuning. See Appendix~\ref{append:how-fine-tuning} for further details.

\subsection{RL/DPO}  
\label{sec:how-rl-dpo}

Reinforcement Learning with Human (or AI) Feedback (RLHF) and Direct Preference Optimization (DPO) enhance user simulators by training policies that maximize rewards or preference scores across multi-turn interactions:
\[
u_t \sim P(C_{t-1}, \Psi_p, \pi_\theta), \quad \pi_\theta = \arg\max_\theta \mathbb{E}[R(\tau)]
\]
where $\pi_\theta$ is a user policy optimized via RL or DPO using feedback from user preferences. 

These methods support adaptive, strategic, and goal-driven behaviors beyond standard supervised learning. Personalization-focused approaches like curiosity-driven RLHF~\cite{wan2025enhancing} reward disambiguation of latent user traits. Memory-aware methods~\cite{seo2024efficient} use DPO to optimize memory selection for factual coherence. ArCHer~\cite{zhou2024archer} introduces hierarchical RL for long-horizon planning via utterance- and token-level control. Offline learning frameworks like hindsight regeneration~\cite{hong2024interactive} revise suboptimal segments using observed user feedback. Action-level DPO~\cite{chen2024learning} improves intent clarification by preferring follow-ups that maximize user satisfaction. See Appendix~\ref{append:how-rl-dpo} for further details.

\begin{table*}[t!]
\vspace{3mm}
\centering
\def\arraystretch{1.2} 
\scriptsize
\renewcommand\BBBBB{\rule[1.45ex]{0pt}{1.45ex}}

\begin{tabular}
{P{0mm}
l
c 
!{\vrule width 0.8pt} 
P{\cellszlg} P{\cellszlg} P{\cellszlg} 
P{\cellszlg} P{\cellszlg}  
!{\vrule width 0.6pt}
P{\cellszlg} 
P{\cellszlg} 
P{\cellszlg}  
P{\cellszlg}  
P{\cellszlg}  
!{\vrule width 0.6pt}
P{\cellszsm} 
P{\cellszsm} 
P{\cellszsm} 
P{\cellszsm} 
P{\cellszsm} 
!{\vrule width 0.6pt}
P{\cellszsm} 
P{\cellszsm} 
P{\cellszsm} 
P{\cellszsm}
P{\cellszsm} 
!{\vrule width 0.8pt}
@{}
}

& & 
& \multicolumn{5}{c!{\vrule width 0.6pt}}{\textcolor{googlegreen}{\textsc{\bfseries\scshape Who}} 
}
& \multicolumn{5}{c!{\vrule width 0.6pt}}{\textcolor{googleblue}{
\textsc{\bfseries\scshape What}}
}
& \multicolumn{5}{c!{\vrule width 0.6pt}}{\textcolor{googlered}{\textsc{\bfseries\scshape How}}} 
& \multicolumn{5}{c!{\vrule width 0.6pt}}{\textcolor{googlepurple}{\textsc{\bfseries\scshape Applications}}} 
\\

& & 
\BBBnew
& \multicolumn{5}{c!{\vrule width 0.6pt}}{(\textbf{Section~\ref{sec:who}})}
\BBBnew
& \multicolumn{5}{c!{\vrule width 0.6pt}}{(\textbf{Section~\ref{sec:what}})}
& \multicolumn{5}{c!{\vrule width 0.6pt}}{(\textbf{Section~\ref{sec:how}})} 
& \multicolumn{5}{c!{\vrule width 0.6pt}}{(\textbf{Section~\ref{sec:apps}})} 
\\

& 
& 
& 
\rotatebox{\rotateDeg}{\textcolor{black}{\textbf{General User Simulation} \textbf{(\S\ref{sec:who-general-users})}}} & 
\rotatebox{\rotateDeg}{\textbf{Persona User Simulation (\S\ref{sec:who-persona-based-sim})}} &
\rotatebox{\rotateDeg}{\textbf{Role Play Simulation (\S\ref{sec:who-role-based-sim})}} &
\rotatebox{\rotateDeg}{\textbf{Individual User (\S\ref{sec:who-user-level-sim})}} &
\rotatebox{\rotateDeg}{\textbf{Hybrid (\S\ref{sec:who-hybrid})}} &
\rotatebox{\rotateDeg}{\textbf{Human-AI (\S\ref{sec:what-human-ai})}} &
\rotatebox{\rotateDeg}{\textbf{Human-Human (\S\ref{sec:what-human-human})}} & 
\rotatebox{\rotateDeg}{\textbf{AI-AI (\S\ref{sec:what-ai-ai})}} &
\rotatebox{\rotateDeg}{\textbf{Many Human-AI (\S\ref{sec:what-many-human-ai})}} &
\rotatebox{\rotateDeg}{\textbf{Hybrid (\S\ref{sec:what-hybrid})}} &
\rotatebox{\rotateDeg}{\textbf{Prompt-based (\S\ref{sec:how-prompt})}} &
\rotatebox{\rotateDeg}{\textbf{RAG (\S\ref{sec:how-rag})}}  &
\rotatebox{\rotateDeg}{\textbf{Fine-Tuning (\S\ref{sec:how-fine-tuning})}} &
\rotatebox{\rotateDeg}{\textbf{RL/DPO (\S\ref{sec:how-rl-dpo})}} &
\rotatebox{\rotateDeg}{\textbf{Hybrid (\S\ref{sec:how-hybrid})}} &
\rotatebox{\rotateDeg}{\textbf{Recommendation}} & 
\rotatebox{\rotateDeg}{\textbf{Summarization}} &
\rotatebox{\rotateDeg}{\textbf{Text Generation}} &
\rotatebox{\rotateDeg}{\textbf{Question Answering}} &
\rotatebox{\rotateDeg}{\textbf{Other}} 
\\
\noalign{\hrule height 0.8pt}

& \sysName{$\mathsf{\sf \bf PersonalConv}$}~\cite{li2025personalized}
&
& {\cellno} 
& {\cellno} 
& {\cellno} 
& {\cellyes} 
& {\cellno} 
& {\cellno} 
& {\cellyes} 
& {\cellno} 
& {\cellno} 
& {\cellno} 
& {\cellyes} 
& {\cellyes} 
& {\cellno} 
& {\cellno} 
& {\cellno} 
& {\cellno} 
& {\cellno} 
& {\cellyes} 
& {\cellno} 
& {\cellyes} 
\\
\hline

& \sysName{$\mathsf{\sf \bf PersonalDialog}$}~\cite{zheng2019personalized}
&
& {\cellno} 
& {\cellno} 
& {\cellno} 
& {\cellyes} 
& {\cellno} 
& {\cellno} 
& {\cellyes} 
& {\cellno} 
& {\cellno} 
& {\cellno} 
& {\cellno} 
& {\cellno} 
& {\cellyes} 
& {\cellno} 
& {\cellno} 
& {\cellno} 
& {\cellno} 
& {\cellyes} 
& {\cellno} 
& {\cellno} 
\\
\hline

& \sysName{$\mathsf{\sf \bf RoleLLM}$}~\cite{wang2023rolellm}
&
& {\cellno} 
& {\cellno} 
& {\cellyes} 
& {\cellno} 
& {\cellno} 
& {\cellyes} 
& {\cellno} 
& {\cellno} 
& {\cellno} 
& {\cellno} 
& {\cellyes} 
& {\cellyes} 
& {\cellyes} 
& {\cellno} 
& {\cellno} 
& {\cellno} 
& {\cellno} 
& {\cellno} 
& {\cellyes} 
& {\cellno} 
\\
\hline

& \sysName{$\mathsf{\sf \bf DMPO}$}~\cite{shi2024direct}
&
& {\cellyes} 
& {\cellno} 
& {\cellno} 
& {\cellno} 
& {\cellno} 
& {\cellyes} 
& {\cellno} 
& {\cellno} 
& {\cellno} 
& {\cellno} 
& {\cellno} 
& {\cellno} 
& {\cellno} 
& {\cellyes} 
& {\cellno} 
& {\cellno} 
& {\cellno} 
& {\cellyes} 
& {\cellno} 
& {\cellno} 
\\
\hline

& \sysName{$\mathsf{\sf \bf SDPO}$}~\cite{kong2025sdpo}
&
& {\cellno} 
& {\cellyes} 
& {\cellno} 
& {\cellno} 
& {\cellno} 
& {\cellno} 
& {\cellyes} 
& {\cellno} 
& {\cellno} 
& {\cellno} 
& {\cellno} 
& {\cellno} 
& {\cellno} 
& {\cellyes} 
& {\cellno} 
& {\cellno} 
& {\cellno} 
& {\cellyes} 
& {\cellno} 
& {\cellno} 
\\
\hline

& \sysName{$\mathsf{\sf \bf Agent Q}$}~\cite{putta2024agent}
&
& {\cellyes} 
& {\cellno} 
& {\cellno} 
& {\cellno} 
& {\cellno} 
& {\cellyes} 
& {\cellno} 
& {\cellno} 
& {\cellno} 
& {\cellno} 
& {\cellno} 
& {\cellno} 
& {\cellno} 
& {\cellyes} 
& {\cellyes} 
& {\cellno} 
& {\cellno} 
& {\cellyes} 
& {\cellyes} 
& {\cellno} 
\\
\hline

& \sysName{$\mathsf{\sf \bf LOOP}$}~\cite{chen2025reinforcement}
&
& {\cellyes} 
& {\cellno} 
& {\cellno} 
& {\cellno} 
& {\cellno} 
& {\cellyes} 
& {\cellno} 
& {\cellno} 
& {\cellno} 
& {\cellno} 
& {\cellno} 
& {\cellno} 
& {\cellno} 
& {\cellyes} 
& {\cellno} 
& {\cellno} 
& {\cellno} 
& {\cellyes} 
& {\cellyes} 
& {\cellno} 
\\
\hline

& \sysName{$\mathsf{\sf \bf KAUCUS}$}~\cite{dhole2024kaucus}
&
& {\cellyes} 
& {\cellno} 
& {\cellno} 
& {\cellno} 
& {\cellno} 
& {\cellyes} 
& {\cellno} 
& {\cellno} 
& {\cellno} 
& {\cellno} 
& {\cellno} 
& {\cellyes} 
& {\cellno} 
& {\cellyes} 
& {\cellyes} 
& {\cellno} 
& {\cellno} 
& {\cellyes} 
& {\cellyes} 
& {\cellno} 
\\
\hline

& \sysName{$\mathsf{\sf \bf CROSS}$}~\cite{yuan2024evaluating}
&
& {\cellno} 
& {\cellno} 
& {\cellyes} 
& {\cellno} 
& {\cellno} 
& {\cellyes} 
& {\cellno} 
& {\cellno} 
& {\cellno} 
& {\cellno} 
& {\cellno} 
& {\cellyes} 
& {\cellno} 
& {\cellno} 
& {\cellyes} 
& {\cellno} 
& {\cellno} 
& {\cellyes} 
& {\cellyes} 
& {\cellno} 
\\
\hline

& \sysName{$\mathsf{\sf \bf PersonaEffect }$}~\cite{hu2024quantifying}
&
& {\cellno} 
& {\cellno} 
& {\cellyes} 
& {\cellno} 
& {\cellno} 
& {\cellyes} 
& {\cellno} 
& {\cellno} 
& {\cellno} 
& {\cellno} 
& {\cellno} 
& {\cellyes} 
& {\cellno} 
& {\cellno} 
& {\cellyes} 
& {\cellno} 
& {\cellno} 
& {\cellyes} 
& {\cellyes} 
& {\cellno} 
\\
\hline

& \sysName{$\mathsf{\sf \bf EvalPersonality }$}~\cite{wang2025user}
&
& {\cellno} 
& {\cellyes} 
& {\cellno} 
& {\cellno} 
& {\cellno} 
& {\cellyes} 
& {\cellno} 
& {\cellno} 
& {\cellno} 
& {\cellno} 
& {\cellyes} 
& {\cellno} 
& {\cellno} 
& {\cellno} 
& {\cellno} 
& {\cellno} 
& {\cellno} 
& {\cellyes} 
& {\cellyes} 
& {\cellno} 
\\
\hline

& \sysName{$\mathsf{\sf \bf P^2 }$}~\cite{jiang2023evaluating}
&
& {\cellno} 
& {\cellyes} 
& {\cellno} 
& {\cellno} 
& {\cellno} 
& {\cellyes} 
& {\cellno} 
& {\cellno} 
& {\cellno} 
& {\cellno} 
& {\cellno} 
& {\cellyes} 
& {\cellno} 
& {\cellno} 
& {\cellno} 
& {\cellno} 
& {\cellno} 
& {\cellno} 
& {\cellyes} 
& {\cellno} 
\\
\hline
& \sysName{$\mathsf{\sf \bf PsyPlay }$}~\cite{yang2025psyplay}
&
& {\cellno} 
& {\cellyes} 
& {\cellyes} 
& {\cellno} 
& {\cellyes} 
& {\cellno} 
& {\cellyes} 
& {\cellno} 
& {\cellno} 
& {\cellno} 
& {\cellyes} 
& {\cellno} 
& {\cellno} 
& {\cellno} 
& {\cellno} 
& {\cellno} 
& {\cellno} 
& {\cellyes} 
& {\cellno} 
& {\cellno} 
\\
\hline

& \sysName{$\mathsf{\sf \bf CSHI }$}~\cite{zhu2025llm}
&
& {\cellyes} 
& {\cellno} 
& {\cellno} 
& {\cellno} 
& {\cellno} 
& {\cellyes} 
& {\cellno} 
& {\cellno} 
& {\cellno} 
& {\cellno} 
& {\cellyes} 
& {\cellyes} 
& {\cellno} 
& {\cellno} 
& {\cellyes} 
& {\cellno} 
& {\cellno} 
& {\cellno} 
& {\cellno} 
& {\cellyes} 
\\
\hline

& \sysName{$\mathsf{\sf \bf RAGate }$}~\cite{wang2024adaptive}
&
& {\cellyes} 
& {\cellno} 
& {\cellno} 
& {\cellno} 
& {\cellno} 
& {\cellyes} 
& {\cellno} 
& {\cellno} 
& {\cellno} 
& {\cellno} 
& {\cellno} 
& {\cellyes} 
& {\cellno} 
& {\cellno} 
& {\cellno} 
& {\cellno} 
& {\cellno} 
& {\cellyes} 
& {\cellyes} 
& {\cellno} 
\\
\hline

& \sysName{$\mathsf{\sf \bf PB\&J }$}~\cite{joshi2025improving}
&
& {\cellno} 
& {\cellyes} 
& {\cellno} 
& {\cellno} 
& {\cellno} 
& {\cellyes} 
& {\cellyes} 
& {\cellno} 
& {\cellno} 
& {\cellyes} 
& {\cellyes} 
& {\cellno} 
& {\cellno} 
& {\cellno} 
& {\cellno} 
& {\cellno} 
& {\cellno} 
& {\cellyes} 
& {\cellyes} 
& {\cellno} 
\\
\hline

& \sysName{$\mathsf{\sf \bf CharacterLLM }$}~\cite{shao2023character}
&
& {\cellno} 
& {\cellno} 
& {\cellyes} 
& {\cellno} 
& {\cellno} 
& {\cellyes} 
& {\cellyes} 
& {\cellno} 
& {\cellno} 
& {\cellno} 
& {\cellno} 
& {\cellyes} 
& {\cellyes} 
& {\cellno} 
& {\cellyes} 
& {\cellno} 
& {\cellno} 
& {\cellyes} 
& {\cellyes} 
& {\cellno} 
\\
\hline

& \sysName{$\mathsf{\sf \bf CharacterBench }$}~\cite{zhou2025characterbench}
&
& {\cellno} 
& {\cellno} 
& {\cellyes} 
& {\cellno} 
& {\cellno} 
& {\cellno} 
& {\cellyes} 
& {\cellno} 
& {\cellno} 
& {\cellno} 
& {\cellno} 
& {\cellyes} 
& {\cellyes} 
& {\cellno} 
& {\cellyes} 
& {\cellno} 
& {\cellno} 
& {\cellno} 
& {\cellyes} 
& {\cellno} 
\\
\hline

& \sysName{$\mathsf{\sf \bf CharacterBench }$}~\cite{huang2024orca}
&
& {\cellno} 
& {\cellyes} 
& {\cellyes} 
& {\cellno} 
& {\cellno} 
& {\cellno} 
& {\cellyes} 
& {\cellno} 
& {\cellno} 
& {\cellno} 
& {\cellno} 
& {\cellyes} 
& {\cellyes} 
& {\cellno} 
& {\cellyes} 
& {\cellno} 
& {\cellno} 
& {\cellno} 
& {\cellyes} 
& {\cellno} 
\\
\hline

& \sysName{$\mathsf{\sf \bf SmallVille }$}~\cite{park2023generative} 
&
& {\cellno} 
& {\cellno} 
& {\cellno} 
& {\cellno} 
& {\cellyes} 
& {\cellno} 
& {\cellyes} 
& {\cellyes} 
& {\cellno} 
& {\cellno} 
& {\cellyes} 
& {\cellno} 
& {\cellno} 
& {\cellno} 
& {\cellno} 
& {\cellno} 
& {\cellno} 
& {\cellno} 
& {\cellno} 
& {\cellyes} 
\\
\hline

& \sysName{$\mathsf{\sf \bf CharMap }$}~\cite{xu2024character} 
&
& {\cellno} 
& {\cellno} 
& {\cellyes} 
& {\cellno} 
& {\cellno} 
& {\cellyes} 
& {\cellyes} 
& {\cellno} 
& {\cellno} 
& {\cellno} 
& {\cellno} 
& {\cellyes} 
& {\cellno} 
& {\cellno} 
& {\cellno} 
& {\cellno} 
& {\cellno} 
& {\cellyes} 
& {\cellno} 
& {\cellno} 
\\
\hline

& \sysName{$\mathsf{\sf \bf DramaLLM }$}~\cite{wu2024role} 
&
& {\cellno} 
& {\cellno} 
& {\cellyes} 
& {\cellno} 
& {\cellyes} 
& {\cellno} 
& {\cellyes} 
& {\cellyes} 
& {\cellno} 
& {\cellno} 
& {\cellno} 
& {\cellno} 
& {\cellyes} 
& {\cellno} 
& {\cellno} 
& {\cellno} 
& {\cellno} 
& {\cellyes} 
& {\cellno} 
& {\cellyes} 
\\
\hline

& \sysName{$\mathsf{\sf \bf LifeStageBench }$}~\cite{fan2025if} 
&
& {\cellno} 
& {\cellno} 
& {\cellyes} 
& {\cellno} 
& {\cellno} 
& {\cellyes} 
& {\cellyes} 
& {\cellno} 
& {\cellno} 
& {\cellno} 
& {\cellno} 
& {\cellno} 
& {\cellno} 
& {\cellno} 
& {\cellno} 
& {\cellno} 
& {\cellno} 
& {\cellyes} 
& {\cellno} 
& {\cellno} 
\\
\hline

& \sysName{$\mathsf{\sf \bf PersonaChat }$}~\cite{zhang2018personalizing} 
&
& {\cellno} 
& {\cellno} 
& {\cellno} 
& {\cellyes} 
& {\cellno} 
& {\cellno} 
& {\cellyes} 
& {\cellno} 
& {\cellno} 
& {\cellno} 
& {\cellyes} 
& {\cellno} 
& {\cellno} 
& {\cellno} 
& {\cellno} 
& {\cellno} 
& {\cellno} 
& {\cellyes} 
& {\cellno} 
& {\cellno} 
\\
\hline

& \sysName{$\mathsf{\sf \bf ProphetChat }$}~\cite{liu2022prophetchat} 
&
& {\cellyes} 
& {\cellno} 
& {\cellno} 
& {\cellno} 
& {\cellno} 
& {\cellyes} 
& {\cellno} 
& {\cellno} 
& {\cellno} 
& {\cellno} 
& {\cellyes} 
& {\cellno} 
& {\cellno} 
& {\cellno} 
& {\cellno} 
& {\cellno} 
& {\cellno} 
& {\cellyes} 
& {\cellyes} 
& {\cellno} 
\\
\hline


& \sysName{$\mathsf{\sf \bf WoW}$}~\cite{dinan2018wizard}
&
& {\cellno} 
& {\cellyes} 
& {\cellno} 
& {\cellno} 
& {\cellno} 
& {\cellno} 
& {\cellyes} 
& {\cellno} 
& {\cellno} 
& {\cellno} 
& {\cellno} 
& {\cellyes} 
& {\cellyes} 
& {\cellno} 
& {\cellno} 
& {\cellno} 
& {\cellno} 
& {\cellyes} 
& {\cellyes} 
& {\cellno} 
\\
\hline

& \sysName{$\mathsf{\sf \bf EmpatheticDialogues}$}~\cite{rashkin2018towards}
&
& {\cellno} 
& {\cellyes} 
& {\cellno} 
& {\cellno} 
& {\cellno} 
& {\cellno} 
& {\cellyes} 
& {\cellno} 
& {\cellno} 
& {\cellno} 
& {\cellno} 
& {\cellno} 
& {\cellyes} 
& {\cellno} 
& {\cellno} 
& {\cellno} 
& {\cellyes} 
& {\cellyes} 
& {\cellno} 
& {\cellyes} 
\\
\hline

& \sysName{$\mathsf{\sf \bf PLATO}$}~\cite{kong2023platolm}
&
& {\cellyes} 
& {\cellno} 
& {\cellno} 
& {\cellno} 
& {\cellno} 
& {\cellno} 
& {\cellyes} 
& {\cellno} 
& {\cellno} 
& {\cellno} 
& {\cellno} 
& {\cellno} 
& {\cellyes} 
& {\cellno} 
& {\cellno} 
& {\cellno} 
& {\cellno} 
& {\cellyes} 
& {\cellno} 
& {\cellno} 
\\
\hline
& \sysName{$\mathsf{\sf \bf ConvEval}$}~\cite{balog2023user}
&
& {\cellyes} 
& {\cellno} 
& {\cellno} 
& {\cellno} 
& {\cellno} 
& {\cellno} 
& {\cellyes} 
& {\cellno} 
& {\cellno} 
& {\cellno} 
& {\cellyes} 
& {\cellyes} 
& {\cellno} 
& {\cellno} 
& {\cellno} 
& {\cellyes} 
& {\cellyes} 
& {\cellno} 
& {\cellno} 
& {\cellno} 
\\
\hline

& \sysName{$\mathsf{\sf \bf UserSimCRS}$}~\cite{afzali2023usersimcrs}
&
& {\cellyes} 
& {\cellno} 
& {\cellno} 
& {\cellno} 
& {\cellno} 
& {\cellno} 
& {\cellyes} 
& {\cellno} 
& {\cellno} 
& {\cellno} 
& {\cellno} 
& {\cellno} 
& {\cellno} 
& {\cellyes} 
& {\cellno} 
& {\cellyes} 
& {\cellyes} 
& {\cellno} 
& {\cellno} 
& {\cellno} 
\\
\hline

\noalign{\hrule height 0.7pt}
\end{tabular}

\vspace{-1mm}
\caption{Overview of the proposed taxonomy for user-simulated data generation techniques and their applications.
Using this taxonomy, we provide a qualitative and quantitative comparison of methods.}

\label{table:qual-and-quant-comparison}
\end{table*}

\subsection{Hybrid Approaches}  
\label{sec:how-hybrid}

Hybrid approaches combine prompting, retrieval, fine-tuning, and RL/DPO to improve realism, controllability, and sample efficiency. Retrieval-augmented fine-tuning (e.g., PRAISE~\cite{kaiser2025preference}) integrates context during training for better grounding. Prompt-to-fine-tune pipelines~\cite{chen2023soulchat} bootstrap data via prompting before supervised tuning. RAG + RL/DPO loops~\cite{wan2025enhancing} coordinate retrieval and policy learning to adaptively query and respond. Hierarchical pipelines like ARCHer~\cite{zhou2024archer} modularize simulation tasks, while personalized stacks~\cite{zhang2024personalization} combine persona prompts, memory, adapters, and RLHF. See Appendix~\ref{append:how-hybrid} for more.

\section{Evaluation} \label{sec:eval}

\paragraph{Traditional Metrics} \label{sec:eval-auto}
Classical metrics like BLEU~\cite{papineni2002bleu}, ROUGE~\cite{lin2004rouge}, slot-F1~\cite{chen2019bert} remain common for structured or goal-oriented dialogues. While efficient and reproducible, these metrics capture narrow facets and are often complemented by human or LLM judges. Human Evaluation remains the gold standard. It is typically conducted through either \emph{interactive} evaluation \emph{offline} evaluation~\cite{ye2021multiwoz, zhou2025characterbench, fan2025if}. For detailed discussion on traditional metrics and human evaluation, see Appendix~\ref{app:traditional-metrics} for details.

\paragraph{LLM-as-Judge}
LLM-as-Judge uses a strong language model as an automatic evaluator, prompted with dialogue context, generated responses, and a rubric targeting dimensions such as coherence, factuality, or safety~\cite{kong2023platolm,fan2025if, lin2023llmevalunifiedmultidimensionalautomatic, saadfalcon2024aresautomatedevaluationframework, zheng2023judgingllmasajudgemtbenchchatbot}. A typical evaluation protocol involves three components: (1) defining evaluation dimensions and rating scales (e.g., 1–5 Likert), (2) providing few-shot exemplars or calibration prompts to align the judge, and (3) instructing the model to explain its reasoning before issuing a final score~\cite{salemi2025reasoningenhancedselftraininglongformpersonalized, li2024llmsasjudgescomprehensivesurveyllmbased, gu2025surveyllmasajudge}. Despite its effectiveness, LLM-as-Judge is sensitive to prompt phrasing and underlying model bias. To address this, recent work proposes symmetric prompting, voting or ensembling across judges, and meta-evaluation by comparing model scores to human ratings~\cite{fan2025if,zhou2025characterbench}.

\paragraph{Trustworthy and Causal Evaluation}
Beyond accuracy-oriented metrics, recent studies emphasize the importance of trustworthy~\cite{liu2024uncertaintyestimationquantificationllms, Ni_Wang_Cheng_Blasch_Derr_2025, zhao2025bewarepomeasuringmitigating} and causal/offline evaluation~\cite{petrov2025llmsestimatingpositionalbias, dudik2014doubly, laban2025llmslostmultiturnconversation} for conversational systems. These paradigms assess not only output quality, but also reliability under uncertainty, robustness to distribution shifts or adversarial prompts, and generalization across topics and user profiles. 
\section{Datasets} \label{sec:datasets}

We categorize commonly used user simulation datasets across dialogue types, with a summary provided in Table~\ref{table:datasets} and additional details in Appendix~\ref{append:datasets}. These datasets span a wide range of interaction settings, including personalized conversations~\cite{zhang2018personalizing, li2025personalized}, multiparty social dialogues~\cite{gao2023livechat}, and information-seeking question answering~\cite{abbasiantaeb2024let, kong2023platolm}. Role-based and character-grounded simulations~\cite{zhou2025characterbench, wang2023rolellm} support more fine-grained persona modeling, while negotiation datasets~\cite{lewis2017deal, he2018decoupling} evaluate goal-driven interaction and strategic reasoning. Other datasets explore long-term memory and context modeling~\cite{fan2025if}, multi-domain generalization~\cite{zhao2018zero}, and dialogue-level response ranking~\cite{zhu2024starling}. This growing corpus of datasets provides a foundation for developing, evaluating, and benchmarking user simulation models under diverse conversational settings.

\section{Applications} \label{sec:apps}
Conversational user simulation underpins a broad spectrum of applications. Before the advent of LLMs, user simulators were already employed for data augmentation in contexts with sparse user histories~\cite{zhao2021action}, and LLMs have further benefited these applications~\cite{zhao2024self}. In conversational recommendation, simulators enable the adaptation of systems to diverse user preferences~\cite{yoon2024evaluating}. In education, they power conversational companions that support interactive learning~\cite{xu2024large}. Last but not the least, in human–computer interaction (HCI), they facilitate tasks such as user interface testing, reducing reliance on human participants~\cite{moore2018conversational}. They have also been applied in specialized domains such as video question answering, where arena-style evaluation with modified Elo rating systems is supported. For further discussion of applications and domains, see Appendix~\ref{app:application}.
\section{Open Problems \& Challenges} \label{sec:open-problems-challenges}

\subsection{Long Conversations}
Simulating extended interactions challenges current models' ability to maintain persona consistency across turns, often leading to drift in style, beliefs, or goals \cite{cho2023crowd,xu2024character}. These issues are amplified in role-based settings, where broken memory or contradictions can lead to hallucinations and character violations \cite{tang2024rolebreak}. Simulated users are also unrealistically cooperative, and long dialogues accumulate errors or lose task focus \cite{putta2024agent}. Solutions require better memory mechanisms, discourse planning, and consistency modeling \cite{kong2025sdpo,fan2025if}.

\subsection{Diversity}
Simulators often mirror cultural-linguistic majorities, yielding overly polite, homogeneous behavior. Though prompting enables personas~\cite{shanahan2023role}, diversity remains limited. Richer simulation requires fine-grained control over traits like emotion, verbosity, and strategy. Most work targets single-user setups, neglecting hybrid or multi-user dynamics essential for realism and personalization.

\subsection{Biases and Toxic Content}
LLM-based simulations risk encoding biases and generating toxic content, especially when personas involve sensitive demographics or public figures \cite{li2025llm,deshpande2023toxicity}. Such biases can harm both research and deployment. While prompt filtering and alignment exist, robust protocols for simulation quality and safety 
are lacking \cite{hu2024quantifying}. For more open problems and challenges, please refer to Appendix~\ref{app:challenges}.
\section{Conclusion} \label{sec:conc}
In this survey, we retrospected the representative literature on LLM-based conversational user simulation through a unified framework along three axes following \textbf{Who}, \textbf{What}, and \textbf{How}. We further provided a broad overview of existing methods used to simulate user conversations, discussed their strengths and limitations, and categorized them across diverse applications. In addition, we examined evaluation protocols and common datasets to support benchmarking. Finally, we outlined key open challenges and suggested future research directions to build more consistent, diverse, and trustworthy user simulators.
\section{Limitations}
While this survey aims to provide a comprehensive overview of LLM-based conversational user simulation, it has several limitations. First, our taxonomy is designed to balance generality and clarity, but certain hybrid or domain-specific methods may not fit perfectly into the proposed categories. Additionally, while we summarize datasets and evaluations, a full benchmarking study across methods was beyond the scope of this work.
\section{Ethical considerations} \label{sec:ethics}
Our work focuses on surveying and organizing existing research on LLM-based conversational user simulation. We aim to support researchers in developing more effective and trustworthy simulation methods. However, although our work is a survey and does not involve human subjects or the creation of new models or datasets, we acknowledge that research on LLM-based conversational user simulation presents meaningful ethical challenges. Role-playing public figures may risk misinformation, reputational harms, or the inappropriate use of likenesses without consent. Similarly, the construction of demographic personas and synthetic dialogues can introduce or reinforce stereotypes, underrepresent certain groups, or produce misleading impressions of lived experiences. The generation of synthetic data further raises questions of provenance, authenticity, and potential downstream misuse, particularly if data are repurposed without clear disclosure.

To mitigate these concerns, we highlight the importance of transparency in documenting the sources and purposes of synthetic data, careful consideration of representational balance when simulating demographic personas, and adherence to community standards around privacy and consent when role-playing individuals or groups. Researchers should also remain mindful of the dual-use risks of simulation technologies, such as their potential application in manipulative or deceptive contexts. By surfacing these issues within our survey, we aim to provide not only an overview of technical progress but also a reminder that the development of user simulation methods must be accompanied by ongoing ethical reflection and responsible practice.

\section*{Acknowledgments}
 This research is supported by Adobe Research and the National Science Foundation (NSF) under grant numbers IIS2239881, IIS2524380, and IIS2524379. 

\bibliography{main}
\bibliographystyle{acl_natbib}

\newpage

\appendix

\section{Background}
\label{sec:background}

\subsection{Simulation in Other Fields}
The foundations of simulation trace back to early philosophical inquiries in Ancient Greece, where understanding processes was a precursor to formal simulation. Since the Industrial Age, simulation has played a central role across scientific disciplines—ranging from physical systems like wind tunnels to computational fluid dynamics. While physics-based simulation, including recent advances in diffusion models, remains well-established, it is not the focus of this work.

Simulation has also been widespread in computer science. For example, the cascade model in social networks~\citep{kempe2003maximizing} simulates how influence spreads through interactions among users. In information retrieval and recommender systems, click models simulate user behavior by modeling how users interact with ranked lists of items~\citep{richardson2007predicting,craswell2008experimental,chuklin2015click}. These models support both system optimization~\citep{kveton2015cascading,lagree2016multiple} and offline evaluation~\citep{joachims2017unbiased,li2018offline}.

Simulation is particularly integral to reinforcement learning (RL), which often requires data-hungry training. Early RL research demonstrated success using handcrafted simulators for controlled domains such as backgammon~\citep{tesauro1994td}, elevator control~\citep{crites1995improving}, and mountain car~\citep{munos1999variable}. Today, standardized simulators like MuJoCo~\citep{todorov2012mujoco} provide high-fidelity environments for physical control tasks. These simulators are grounded in well-understood dynamics, unlike human behavior, which remains far more complex and less predictable. Building large-scale, diverse human simulators remains a core challenge.

Inspired by RL success in physical environments, researchers have also developed simulators for human-centric domains such as recommender systems~\citep{rohde2018recogym,ie2019recsim}. However, most of these frameworks require learning user simulators from domain-specific data, a task complicated by weak sequential signals in common datasets like MovieLens~\citep{harper2015movielens}, where preference patterns often reflect static biases rather than temporal interactions~\citep{koren2009matrix}.

In statistics, simulation also plays a role in counterfactual reasoning, which estimates treatment effects from logged data without costly or infeasible A/B tests~\citep{bottou2013counterfactual}. Matching methods~\citep{stuart10matching} attempt to estimate treatment effects by pairing treated and untreated samples with similar covariates, though imbalance often introduces bias. Propensity-based estimators—such as inverse propensity scoring~\citep{horvitz1952generalization}, clipped estimators~\citep{ionides2008truncated}, and doubly robust approaches~\citep{robins1994estimation,dudik2014doubly}—face similar issues. In such settings, simulation holds promise as a way to impute missing data or generate counterfactual outcomes, including for complex objects like conversations.

\subsection{Human Modeling}

Modeling of human preferences has been studied extensively. Pairwise preferences
are often modeled using the Bradley-Terry-Luce model~\citep{bradley1952rank}.
Permutation preferences are frequently modeled using the Plackett-Luce model~\citep{plackett1975analysis}. These models are commonly known as discrete choice
models~\citep{train2009discrete}. Modeling and learning from human feedback has
recently found many applications in machine learning~\citep{christiano2017deep},
notably in reinforcement learning from human feedback (RLHF)~\citep{ouyang2022training} and direct preference optimization (DPO)~\citep{rafailov2023direct}. Generative human simulation take these efforts to the
next level, by modeling complex human interactions with humans and not just relative preferences.

\subsection{Comparing to Existing Surveys}
\label{app:survey_diff}
Recently, related surveys have been published in the domain. However, we want to point out that our surveys differs in multiple aspects comparing to existing surveys, as shown in Table~\ref{tab:survey_comparison}. First, instead of focusing on a simple simulation target~\cite{chen2024persona}, we provide a comprehensive analysis on the variety of targets in the who section. Additionally, although data generation is a key conversational application~\cite{soudani2024surveyrecentadvancesconversational}, we do not limit our discussion to data generation by including extended discussions on various forms of simulations such as individual simulation. Last but not least, we focus exclusively on conversational data, which enables us to provide a more detailed and systematic discussion on conversation specific topics and techniques such as conversation paradigms comparing to prior works~\cite{zhang2024personalization, balog2025user}.

\begin{table*}[t]
\centering
\scriptsize
\caption{Comparison of Existing Surveys and Our Survey on LLM-based Conversational User Simulation.
Symbols indicate coverage level: $\bullet$ = covered; $\circ$ = not covered; $\circledcirc$ = partially covered.}
\label{tab:survey_comparison}
\begin{tabular}{lccccccc}
\toprule
\textbf{Survey} 
& \textbf{Who} 
& \textbf{Who} 
& \textbf{Who} 
& \textbf{What} 
& \textbf{How} 
& \textbf{Conv.} 
& \textbf{Applications} \\
& \textbf{(User-level)} 
& \textbf{(Persona / Role-play)} 
& \textbf{(Global / Default)} 
& \textbf{(Conv. Paradigm)} 
& \textbf{(Techniques)} 
& \textbf{Evaluation} 
&  \\
\midrule
\citet{zhang2024personalization} 
& $\bullet$ & $\circledcirc$ & $\circ$ & $\circ$ & $\bullet$ & $\circ$ & $\circledcirc$ \\

\citet{balog2025user} 
& $\circledcirc$ & $\circledcirc$ & $\circ$ & $\circ$ & $\bullet$ & $\circ$ & $\circledcirc$ \\

\citet{soudani2024surveyrecentadvancesconversational} 
& $\circ$ & $\circ$ & $\bullet$ & $\circledcirc$ & $\circ$ & $\circ$ & $\bullet$ \\

\citet{chen2024persona} 
& $\circ$ & $\bullet$ & $\circ$ & $\circledcirc$ & $\circledcirc$ & $\circledcirc$ & $\bullet$ \\

\midrule
\textbf{Ours} 
& $\bullet$ & $\bullet$ & $\bullet$ & $\bullet$ & $\bullet$ & $\bullet$ & $\bullet$ \\
\bottomrule
\end{tabular}
\end{table*}

\definecolor{googleblue}{HTML}{4285F4}
\definecolor{googlered}{HTML}{DB4437}
\definecolor{googlepurple}{HTML}{A142F4} 
\definecolor{googlegreen}{HTML}{0F9D58}

\begin{table*}[t!]
\centering
\small
\renewcommand{\arraystretch}{1.30} 
\begin{tabularx}{1.0\linewidth}{
>{\centering\arraybackslash}m{15mm} 
>{\RaggedLeft\arraybackslash}m{55mm} 
X
}
\toprule

\textbf{Taxonomy} &
\textbf{Category} & 
\textbf{Description}
\\ 
\midrule

\multirow{5}{*}{\textcolor{googlegreen}{\textbf{\makecell{Who\\\textcolor{black}{(\textbf{Section~\ref{sec:who}})}}}}} 
& General User (Sec.~\ref{sec:who-general-users}) &   
This class of techniques performs simulation of a general user conversation. This is the most general class of techniques.
\\

& Persona-level (Sec.~\ref{sec:who-persona-based-sim}) &  
This class of techniques leverages demographics and other persona-based data to simulate conversations

\\

& Role Play (Sec.~\ref{sec:who-role-based-sim}) &  
The class of techniques leverages role playing to simulate conversations based on specific 

\\

& Individual User (Sec.~\ref{sec:who-user-level-sim}) &  
The class of conversational simulation techniques focus on generating specific conversations and other facets tailored to a specific user.
\\

& Hybrid (Sec.~\ref{sec:who-hybrid}) &  

This class of techniques combines techniques from two or more of the prior categories of techniques.
\\

\midrule

\multirow{5}{*}{\textcolor{googleblue}{\textbf{\makecell{What\\\textcolor{black}{(\textbf{Section~\ref{sec:what}})}}}}}
& Human-AI (Sec.~\ref{sec:what-human-ai}) &  
This class of conversation objective simulates the interaction between human and AI. Commonly used for instruction-finetuning.

\\

& Human-Human (Sec.~\ref{sec:what-human-human}) & 
This class of conversation objective focuses on simulating the conversation between human, often associated with assigned personalities.

\\

& AI-AI
(Sec.~\ref{sec:what-ai-ai}) & 
This class of conversation objective revolves around the open-ended interaction between two or more AI agents.
\\

& Many Human-AI
(Sec.~\ref{sec:what-many-human-ai}) & 
This class of conversation objective explores the collaborative conversation trajectories between multiple human and AI. 
\\

& Hybrid (Sec.~\ref{sec:what-hybrid}) &  
This class of conversation objective combines two or more of the above conversational simulation paradigms. 

\\

\midrule

\multirow{5}{*}{\textcolor{googlered}{\textbf{\makecell{How~\\\textcolor{black}{(\textbf{Section~\ref{sec:how}})}}}}}
& Prompt-based (Sec.~\ref{sec:how-prompt}) &  
This class of techniques directly uses prompts to steer LLMs toward simulating user conversations.
\\

& RAG (Sec.~\ref{sec:how-rag}) &  
Retrieval-Augmented Generation methods incorporate external knowledge into simulation by retrieving relevant context.
\\

& Fine-Tuning (Sec.~\ref{sec:how-fine-tuning}) &  
These techniques adapt LLMs to user simulation tasks through supervised training (e.g. LoRA) on dialogue data.
\\

& RL/DPO (Sec.~\ref{sec:how-rl-dpo}) &  
This class trains user simulators using feedback-driven optimization such as Reinforcement Learning and Direct Preference Optimization.
\\

& Hybrid (Sec.~\ref{sec:how-hybrid}) &  
Hybrid methods combine prompting, retrieval, fine-tuning, and RL/DPO, often using modular or multi-stage architectures.
\\
\bottomrule
\end{tabularx}
\caption{Taxonomy of User Conversation Simulation (Section~\ref{sec:who}-\ref{sec:how}).}
\label{tab:taxonomy-techniques}
\end{table*}

\begin{table}[t]
    \centering
    \small
    \caption{Taxonomy of General User Simulation}
    \label{tab:taxonomy_general_user}
    \begin{tabularx}{0.49\textwidth}{@{}ll@{}}
        \toprule
        \textbf{Category} &
        \textbf{Work(s)} \\
        \midrule
        \multirow{3}{*}{\textbf{Trajectory Optimization}} &
        M-DPO~\cite{xiong2024building}, \\
        & SDPO~\cite{kong2025sdpo}, \\
        & LOOP~\cite{chen2025reinforcement} \\
        \midrule
        
        \multirow{2}{*}{\textbf{Exploration Optimization}} &
        ETO~\cite{song2024trial}, \\
        & AgentQ~\cite{putta2024agent} \\
        \midrule
        
        \textbf{Diverse User Simulation} &
        KAUCuS~\cite{dhole2024kaucus} \\
        \bottomrule
    \end{tabularx}
    \vspace{-0.15in}
\end{table}

\section{Who: Simulation Target (Extended)}
\subsection{General User Simulation}
\label{append:general-user}
To optimize customized dialogue systems for general user interactions, recent work has focused on improving conversational performance at the turn level and trajectory level. M-DPO~\cite{xiong2024building} introduces a multi-turn online iterative framework for direct preference learning, specifically designed to handle multi-turn reasoning and tool integration. Building on trajectory-based optimization, ETO~\cite{song2024trial} develops an exploration-based approach that learns from past exploration trajectories, including failure cases, to improve performance. Additionally, to address training noise in long conversations, SDPO~\cite{kong2025sdpo} focuses on leveraging specific meaningful segments within conversations while minimizing the impact of less relevant interactions. AgentQ~\cite{putta2024agent} further improves the trajectory exploration by addressing the sub-optimal policy outcomes due to compounding errors and limited exploration data. It combines Monte Carlo Tree Search (MCTS) with self-critique and iterative fine-tuning, learning from both positive and negative conversational trajectories.

More recently, LOOP~\cite{chen2025reinforcement} trains interactive assistants directly in their target environments by formulating the training as a partially observable Markov decision process. This approach uses a data and memory-efficient variant of Proximal Policy Optimization (PPO) that eliminates the need for value networks and requires only a single copy of the language model in memory. Complementing these optimization approaches, work by \citet{dhole2024kaucus} focuses on generating diverse user simulations that capture varied interaction patterns between users and assistants. These simulations provide training data for developing more helpful and robust conversational agents that can handle the breadth of general user interactions.

\begin{table}[t]
    \centering
    \small
    \caption{Taxonomy of Persona-level User Simulation.}
    \label{tab:taxonomy_persona_user}
    \begin{tabularx}{0.45\textwidth}{@{}ll@{}}
        \toprule
        \textbf{Category} &
        \textbf{Work(s)} \\
        \midrule
        \multirow{2}{*}{\makecell[l]{\textbf{Descriptive} \\ \textbf{Features}}}
        & Prompting~\cite{hu2024quantifying}, \\
        & PsyPlay~\cite{yang2025psyplay} \\
        \midrule
        
        \multirow{3}{*}{\makecell[l]{\textbf{Personality} \\ \textbf{Framework}}}
        & Orca~\cite{huang2024orca},\\
        & HEXACO~\cite{ji2024persona}, \\
        & PB\&J~\cite{joshi2025improving} \\
        \midrule
        
        \multirow{2}{*}{\textbf{Benchmarks}}
        & PersonaCatch~\cite{li2025llm},\\
        & PersonaEval~\cite{wang2025user} \\
        \bottomrule
    \end{tabularx}
\end{table}

\subsection{Persona-level User Simulation}
\label{append:personal-simulation}
Persona-level simulation primarily relies on explicitly defining user characteristics for LLMs. Early approaches use direct prompting with descriptive attributes (e.g., demographics)~\cite{hu2024quantifying, Qiu_2024}, while more recent work grounds simulation in psycho-social theory by incorporating established personality frameworks such as the Big Five~\cite{wang2025user, jiang2023evaluating, li2025llm, yang2025psyplay, jiang2023personallm, serapio2023personality} and HEXACO~\cite{ji2024persona}. These efforts enable LLMs to simulate users with more realistic and consistent social psychometrics. Frameworks like Orca~\cite{huang2024orca} augment generation with personal context, and PsyPlay~\cite{yang2025psyplay} builds personality-infused agents capable of portraying designated traits. Beyond persona, more recent work~\cite{mehri2025goalalignmentllmbaseduser} focuses on aligning the goals of the simulated persona.

Beyond prompting, fine-tuning methods such as Supervised Fine-Tuning (SFT) and Direct Preference Optimization (DPO)~\cite{li2025llm} offer greater control, while techniques like Personality Activation Search (PAS)~\cite{zhu2024personality} adjust internal activations to align with group preferences. These approaches demonstrate that LLM personas can consistently reflect assigned traits in psychometric tests~\cite{wang2025user} and exhibit representative linguistic patterns~\cite{jiang2023personallm}.

However, a critical concern is that persona assignment may introduce systematic biases, leading to skewed outcomes such as political forecasts~\cite{li2025llm}, or increased toxicity~\cite{deshpande2023toxicity}, potentially reinforcing harmful stereotypes. These limitations underscore the need for better validation~\cite{hu2024quantifying} and more trustworthy simulation practices.

\begin{table}[t]
    \centering
    \small
    \caption{Taxonomy of \emph{Role Play Simulation}.}
    \label{tab:taxonomy_role_play}
    \begin{tabularx}{0.46\textwidth}{@{}ll@{}}
        \toprule
        \textbf{Category} &
        \textbf{Method(s)} \\
        \midrule
        
        \multirow{5}{*}{\textbf{Frameworks}}
        & Ditto~\cite{lu2024large}, \\
        & CharacterLLM~\cite{shao2023character} \\
        & PCL~\cite{ji2025enhancing} \\
        & RoleCraft~\cite{tao2023rolecraft} \\
        & CharMap~\cite{xu2024character} \\ 
        \midrule
        
        \multirow{3}{*}{\textbf{\makecell[l]{Evaluation \\ Benchmarks}}}
        & RoleBench~\cite{wang2023rolellm},\\
        & CharacterBench~\cite{zhou2025characterbench},\\
        & PsyPlay-Bench~\cite{yang2025psyplay} \\
        \midrule
        
        \multirow{2}{*}{\textbf{Challenges \& Risks}}
        & RoleBreak~\cite{tang2024rolebreak},\\
        & Toxicity~\cite{deshpande2023toxicity}\\
        
        \bottomrule
    \end{tabularx}
\end{table}

\subsection{Role Play Simulation}
\label{append:role-play-simulation}
Role play simulation is fundamentally defined by the implicit knowledge of language models~\cite{serapio2023personality, lu2025agentrewardbench}, which allows LLMs to inherently associate the role with a rich set of personal traits and psychosocial status~\cite{wang2023rolellm}. The capability of LLMs to capture nuanced relationships within extensive textual data means they can simulate more diverse and complex behaviors, reflecting the variability of human decision-making~\cite{wang2025user}.

To elicit and shape this implicit knowledge, various techniques are proposed. Direct ``role-play prompting'' instructs LLMs to generate like a specific character, leveraging their steer-ability to tailor output style and tone~\cite{wang2025user, wang2023rolellm}. More advanced methods focus on training LLMs as ``trainable agents'' specifically for role-playing, improving consistency and naturalness through fine-tuning on character-related dialogues~\cite{shao2023character, lu2024large, tao2023rolecraft}. More recently, ~\citet{ji2025enhancing} demonstrates that contrastive learning through adversarial self-play can enhance character consistency and prevent deviations from the assigned character. ~\citet{qiu-lan-2025-psydial} further shows that letting the role play agent to reconstruct the masked dialogue can improve the generation quality.

Leveraging the implicit character knowledge allows role-playing agents to make persona-driven decisions that align with their assigned identities~\cite{xu2024character, park2023generative, qiu2024interactiveagentssimulatingcounselorclient}. This capability moves role-play beyond mere conversation to influence choices in various scenarios, with character-level benchmarks introduced to evaluate decision-making consistency~\cite{xu2024character}. Role-playing can enable complex ``drama-interaction'' systems where multiple LLMs simulate intricate social dynamics~\cite{wu2024role} or act as interactive simulacra of human behavior in social science research~\cite{wang2025user, park2023generative}. The ability for an LLM to ``know its own story'' as a character, including its past and experiences, is an emerging area of research exploring lifelong learning in LLMs~\cite{fan2025if}.

The reliance on implicit knowledge also introduces significant challenges and risks. A primary concern is \textit{character hallucination}, where LLMs generate content inconsistent with the assigned character's known attribute. For example, asking a Mozart character programming problems can lead to unexpected output. Recent research has shown that such hallucination can be exploited as jailbreak attacks~\cite{tang2024rolebreak,
sadeq2024mitigating}. In addition, pre-existing biases tied to
famous figures or demographics in the training data can lead to increased toxicity and the propagation of incorrect stereotypes when LLMs role-play specific individuals, which can result in outputs that are defamatory or harmful~\cite{deshpande2023toxicity}.
Furthermore, the robustness of role-playing can decrease as the complexity of the roles increases~\cite{wang2025user}, and LLMs might exhibit \textit{persona collapse} or low consistency in challenging situations~\cite{xu2024character}. 

\subsubsection{Evaluation on Role-Playing Benchmarks}
\label{app:roleplay-eval}

This subsection summarizes representative evaluation results on commonly used role-playing benchmarks, illustrating how different modeling paradigms perform in character simulation and role consistency.

\paragraph{RoleLLM.}
Table~\ref{tab:rolellm} below reports results on RoleLLM~\cite{wang2023rolellm}, where CUS measures character understanding, RAW evaluates response appropriateness, and SPE assesses role-specific knowledge. Prompt-based large models achieve strong overall performance, while fine-tuned models—despite smaller model sizes—show competitive results and notably stronger role knowledge (SPE).

\begin{table}[h]
\centering
\scriptsize
\caption{Evaluation on RoleLLM~\cite{wang2023rolellm}.}
\label{tab:rolellm}
\begin{tabular}{lcccc}
\toprule
\textbf{Model} & \textbf{CUS} & \textbf{RAW} & \textbf{SPE} & \textbf{Avg.} \\
\midrule
RoleGPT (Prompt-Based) & 57.6 & 53.2 & 32.3 & 47.7 \\
ChatPLUG (RAG w/ LLaMA) & 24.0 & 34.7 & 25.8 & 28.2 \\
Character.AI (Prompt-Based LLaMA) & 41.9 & 45.7 & 30.3 & 39.3 \\
RoleLLaMA-7B (Fine-tuned) & 32.9 & 37.6 & 38.1 & 36.2 \\
RoleLLaMA2-13B (Fine-tuned) & 37.5 & 47.9 & 48.8 & 44.7 \\
\bottomrule
\end{tabular}
\end{table}

\paragraph{WikiRole.}
Table~\ref{tab:wikirole} presents evaluation on WikiRole~\cite{dinan2018wizard}, focusing on role accuracy and explicit role knowledge. While large prompt-based models perform well overall, fine-tuned models show clear gains in role knowledge.

\begin{table}[h]
\centering
\scriptsize
\caption{Evaluation on WikiRole~\cite{dinan2018wizard}.}
\label{tab:wikirole}
\begin{tabular}{lcc}
\toprule
\textbf{Model} & \textbf{Accuracy} & \textbf{Role Knowledge} \\
\midrule
GPT-4 (Prompt-Based) & 80.0 & 76.2 \\
CharacterGLM (Self-Play Fine-tuned) & 75.0 & 47.3 \\
Qwen-72B + Chat & 90.0 & 66.4 \\
\bottomrule
\end{tabular}
\end{table}

\paragraph{RoleInstruct.}
Table~\ref{tab:roleinstruct} reports results on RoleInstruct~\cite{tao2023rolecraft}, where fine-tuned models consistently outperform prompt-only baselines on role-specific knowledge (SPE).

\begin{table}[h]
\centering
\scriptsize
\caption{Evaluation on RoleInstruct~\cite{tao2023rolecraft}.}
\label{tab:roleinstruct}
\begin{tabular}{lcccc}
\toprule
\textbf{Model} & \textbf{CUS} & \textbf{RAW} & \textbf{SPE} & \textbf{Avg.} \\
\midrule
GPT-4 (Prompt-Based) & 52.0 & 55.7 & 28.3 & 45.3 \\
RoleGLM (Fine-tuned) & 50.5 & 52.6 & 34.0 & 45.7 \\
RoleCraft-GLM (Self-Play Fine-tuned) & 51.5 & 53.8 & 35.7 & 47.0 \\
\bottomrule
\end{tabular}
\end{table}

\paragraph{Summary.}
Across role-playing benchmarks, large prompt-based models benefit from scale and implicit knowledge, achieving strong baseline performance. However, fine-tuned models—often built on smaller backbones—consistently improve role-specific knowledge and character consistency, particularly on SPE metrics. In contrast, retrieval-augmented approaches such as ChatPLUG generally underperform, suggesting that fine-tuning is more effective than RAG for role simulation and character elicitation.

\subsection{Individual User Simulation}
\label{append:individual-simulation}
The most straightforward individual user simulation paradigm is through the integration of user profiles into prompts. For example, PERSONA-CHAT~\cite{zhang2018personalizing} provides crowdworkers with multi-sentence user profiles, enabling individualized realistic user simulation. The FoCus dataset~\cite{jang2022call} extends this idea by incorporating Wikipedia-based facts into persona grounding. These approaches enhance the response by aligning responses with a user's stated context.

In order to incorporate more nuanced individual profiles, another line of research leverages a user's \textit{dialogue history} to learn communication style and evolving preferences. PHMN~\cite{li2021dialogue} extracts long-term dialogue patterns such as personalized word usage and context attention, improving multi-turn response selection. 

To handle long-range personalization, \textit{multi-session memory} mechanisms have been introduced. The MSC dataset~\cite{xu2021beyond} captures multiple-session chats with user-specific memory summaries to enable coherent re-engagement. Mem0~\cite{chhikara2025mem0} proposes a production-ready memory architecture that dynamically extracts and consolidates user memories across sessions. 

Finally, more recent efforts collect \textit{real-world personality traits} to ground simulation in natural behavior. The RealPersonaChat dataset~\cite{yamashita2023realpersonachat} uses real personality scores (e.g., Big Five traits) in free-form Japanese chats, helping the model reflect actual personality cues. LiveChat~\cite{gao2023livechat} mines detailed persona profiles and interactions from live-streaming platforms, constructing a large-scale Chinese corpus with naturally occurring individual variation.

\begin{figure}
    \centering
    \includegraphics[width=0.75\linewidth]{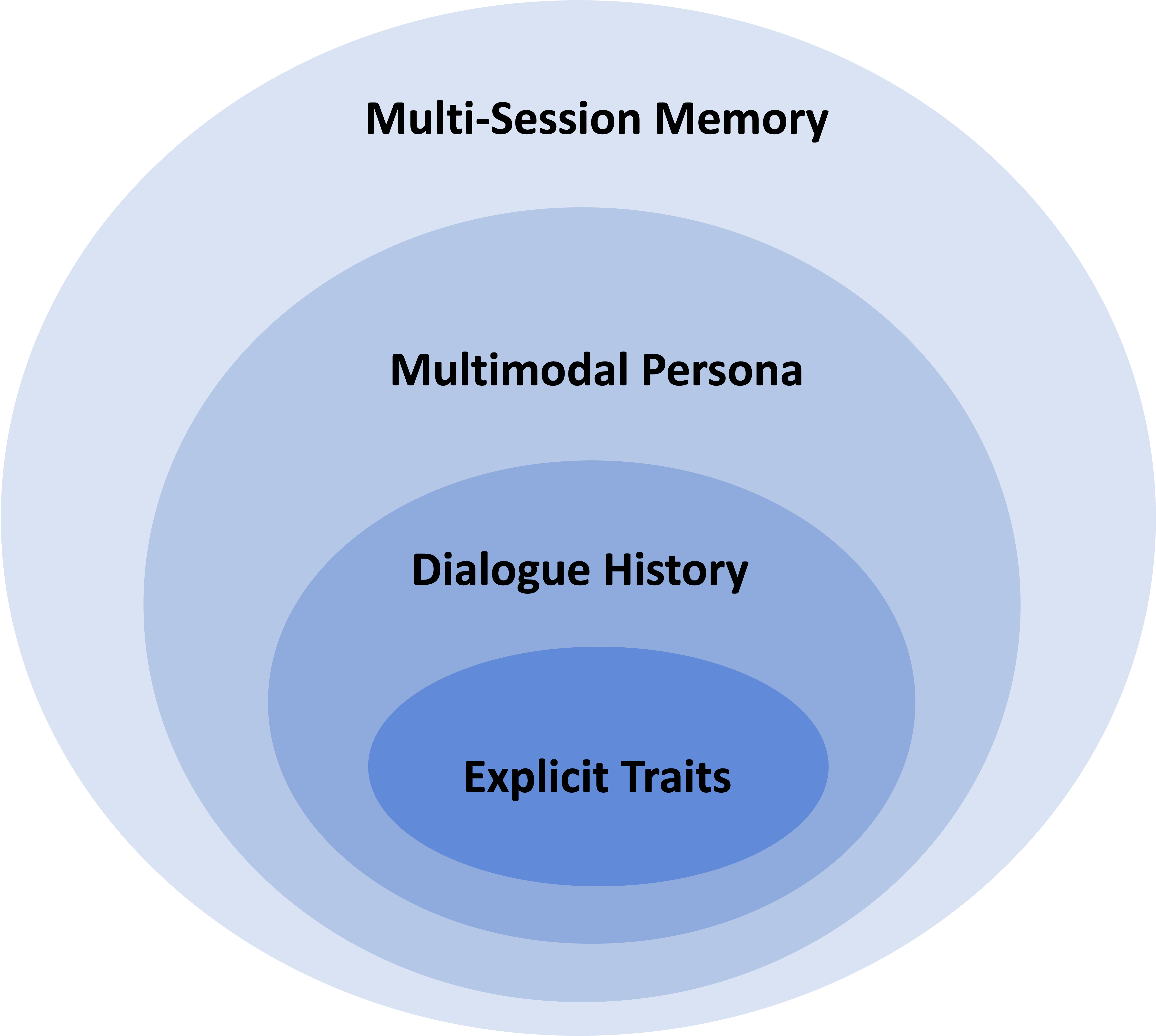}
    \caption{Taxonomy of Individual User Simulation.
    Explicit traits
    }
    \label{fig:individual_user_taxonomy}
\end{figure}

\begin{figure*}
    \centering
    \subfigure[Human-Human Conversation Trajectories]{
    \includegraphics[width=0.45\linewidth]{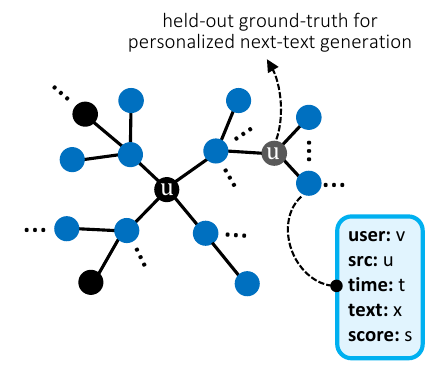}
    }
    \subfigure[Human-AI Conversation Trajectories]{
    \includegraphics[width=0.45\linewidth]{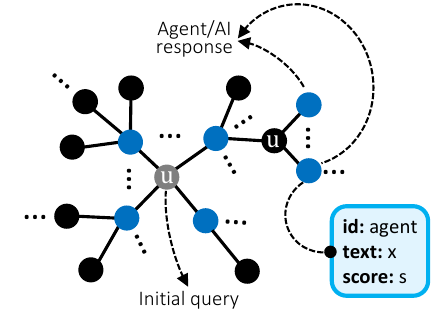}
    }
    \caption{%
    Comparing different types of conversational trajectories starting from an initial input query by the user.
    }
    \label{fig:conv-taxonomy}
\end{figure*}

\section{What: Simulation Objectives (Extended)}
\subsection{Human-AI Simulation}
\label{append:human-ai}
Generating such conversations has attracted significant attention due to the scarcity of high-quality, diverse human-annotated data and the high cost of large-scale annotation~\cite{wang2022self, xu2024wizardlm, li2024synthetic}. Recent studies have proposed various frameworks for synthesizing instruction-following datasets. One prominent approach focuses on bootstrapping instruction datasets through self-generation. The self-instruct framework~\cite{wang2022self} introduces a semi-automated pipeline in which a pretrained LLM generates synthetic instructions and input-output pairs. These synthetic examples are filtered and then used to fine-tune the model itself, improving its instruction-following capabilities. Building upon this idea, WizardLM~\cite{xu2024wizardlm} progressively rewrite simple instructions into more complex forms, enabling LLMs to handle multi-step reasoning and diverse task demands. To further reduce reliance on heuristic design, Auto Evol-Instruct~\cite{zeng2024autoevol} fully automates the evolution process, iteratively optimizing instruction datasets without any human intervention.

Another line of work emphasizes generalized and domain-agnostic data generation. GLAN~\cite{li2024synthetic} introduces a paradigm inspired by educational taxonomies, decomposing human knowledge into fields to generate instructions spanning a wide range of domains. Building on this idea in multimodal contexts, Multimodal Self-Instruct~\cite{zhang2024multimodal} synthesizes visual instruction reasoning tasks using abstract images. ProphetChat~\cite{liu2022prophetchat} further extends this direction by simulating future conversations based on existing dialogue histories.

A complementary strategy focuses on task-specific and tailored data alignment. Recognizing that different downstream applications require distinct instruction distributions, CodecLM~\cite{wang2024codeclm} introduces a metadata-based Encode-Decode process to generate high-quality synthetic data tailored for specific tasks. In parallel, \citet{chan2024balancing} empirically evaluate trade-offs in synthetic data generation strategies under various resource constraints, providing practical guidelines for selecting the augmentation methods. Additionally, DataDreamer~\cite{patel2024datadreamer} provides a unified library for chaining synthetic data generation, fine-tuning, and evaluation, emphasizing reproducibility and practical pipeline.

In addition, a growing body of work explores domain-specific human–machine simulations tailored to specialized applications. For instance, synthetic patient–physician dialogues are generated to support medical AI systems~\cite{das2024synthetic}, while museum guidance agents are developed through curated human–AI conversations~\cite{rachidi2025design}. These efforts highlight the importance of aligning simulation data with the nuanced requirements of different domains.

\subsection{Human-Human Simulation}
\label{append:human-human}
Early sequence-to-sequence chatbots demonstrated fluency but often produced generic responses and lacked consistent speaker personalities~\cite{vinyals2015neural}. To address this, PersonaChat~\cite{zhang2018personalizing} provide each speaker with a personal profile of facts, encouraging models to maintain a coherent identity throughout the conversation. Beyond persona consistency, other works injected additional context to make dialogues more natural. Wizard-of-Wikipedia dataset~\cite{dinan2018wizard} tasks one participant (the ``wizard'') with retrieving and using Wikipedia facts during conversation. Models trained on these dialogues learn to incorporate factual knowledge into their responses, enabling more informative and substantive discussions. The EmpatheticDialogues dataset~\cite{rashkin2018towards}, consisting of 25k human–human conversations, was collected to teach models how to recognize and respond to a conversation partner's emotions. Systems fine-tuned on these emotionally grounded dialogues have been shown to be perceived as significantly more empathetic by human evaluators, compared to models trained only on general chat data.  

In \emph{task-oriented dialogue} domains (e.g., booking tickets), human–human simulation is equally important. Large-scale corpora such as MultiWOZ~\cite{budzianowski2018multiwoz} illustrate the collection of human–human written conversations spanning multiple domains for developing task-oriented agents. MultiWOZ contains around 10k dialogues collected via crowd workers role-playing as user and assistant. While such Wizard-of-Oz style data provide valuable training and evaluation scenarios, creating a new human-human dataset for every domain is time-consuming and expensive.  

To mitigate the problem of data scarcity, researchers have explored \emph{synthetic dialogue generation}. ~\citet{shah2018bootstrapping} proposed a self-play bootstrapping approach for task-oriented dialogues: using a predefined domain schema and slot-filling logic, a simulated ``user'' and ``agent'' were auto-generated to produce dialogues, which were then paraphrased and validated by crowd workers. This method produced additional training trajectories without relying entirely on manually written dialogues, effectively augmenting the data for learning dialog policies.

\subsection{AI-AI Simulation}
\label{append:AI-AI}
One primary use case of AI–AI simulation is the study of \emph{emergent behavior and social dynamics} by placing multiple LLM-driven agents in shared virtual environments and allowing them to interact over extended time horizons~\cite{haase2025beyond}. Rather than optimizing for a specific task, these simulations aim to observe how complex interaction patterns arise from simple initial conditions and agent profiles. For example, Park et al.~\cite{park2023generative} instantiated a small community of generative agents, each equipped with a unique backstory, memory module, and behavioral script. These agents autonomously engaged in open-ended conversations, formed social relationships, and coordinated group activities—exhibiting behavior reminiscent of real-world communities. CRSEC~\cite{ren2024emergence} further examined how repeated local interactions among agents lead to the emergence of shared social norms and behavioral conventions over time. Similarly, AgentSociety~\cite{piao2025agentsociety} demonstrated that large-scale LLM-based simulations can replicate macro-level social patterns and align with real-world trajectories of community evolution. Such results suggest that today's large language models are not only capable of coherent multi-agent dialogue but can also simulate realistic, temporally grounded social processes. Consequently, AI–AI simulation has emerged as a powerful tool for generating rich interaction data and conducting behavioral evaluations at scale, offering a scalable and cost-effective alternative to traditional human-annotated datasets.

Beyond data augmentation, AI–AI simulation has become a framework for \emph{multi-agent cooperation}~\cite{guo2024large}. CAMEL~\cite{li2023camel} uses a role-playing paradigm - one agent is assigned as the ``user'' (task setter) and another as the ``assistant'' (solution provider) - to autonomously interact to complete a given task. Similarly, AutoGen~\cite{wu2023autogen} defines multiple conversable agents (LLMs, tools, or human proxies) that communicate with each other to complete complex tasks. These frameworks demonstrated that solutions obtained via these AI–AI conversations can outperform those from a single agent working alone. 

While many AI–AI simulations focus on collaboration, others explore the \emph{adversarial} dynamics among agents. For instance, ~\citet{du2023improving} demonstrated that adversarial debate between LLMs can lead to improvements in both factuality and reasoning, as the models iteratively challenge and refine each other's claims. Furthermore, ~\citet{rennard2025bias} introduced a self-debate framework in which two instances of a language model argue opposing viewpoints to persuade a third, neutral LLM judge. This setup enables the examination of how persuasive strategies and biases propagate across models. More recently, ~\citet{hua2024game} studied adversarial behavior in game-theoretic scenarios, revealing that unstructured LLMs often deviate from rational strategies. To mitigate this, they introduced \emph{agentic workflows} that guide LLM reasoning through structured decision-making procedures, enabling more consistent alignment with game-theory principles. 

\subsection{Many Human-AI Simulation}
\label{append:many-human}
Historically, most research on human–AI collaboration has centered on single-user scenarios, where the AI functions as a personal assistant or tutor. However, there is growing recognition that future AI agents will need to operate as members of teams rather than merely serve individual users in isolation. Recent work has begun to explore this paradigm shift. For example, ChatCollab~\cite{klieger2024chatcollab} treats the AI agent as a peer collaborator within a team-based software engineering setting. The system enables joint problem-solving among multiple human participants and one or more AI agents, and the results suggest that AI can occupy differentiated roles within human teams—such as idea contributor, critic, or coordinator—mirroring the diversity of human collaboration styles.

Additionally, Dittos~\cite{leong2024dittos} introduces a personalized AI agent capable of standing in for a human teammate during remote meetings. These embodied agents are not merely passive note-takers but are designed to actively participate on behalf of the absent user. Experiments show that such agents can evoke feelings of presence and trust among human team members, even contributing meaningfully to group discussions. Furthermore, the Multi-User Chat Assistant (MUCA)~\cite{mao2024multi} framework introduces a simulation-based approach to multi-party dialogue. MUCA employs a multi-user simulator to mimic the behaviors of several distinct human participants, enabling the training and evaluation of group-aware AI assistants. By modeling not just individual utterances but the evolving group dynamics over time, MUCA facilitates the development of more socially intelligent AI agents capable of navigating complex multi-user settings.

Despite these promising directions, research on many-human–AI simulation remains relatively underexplored. Most existing systems are still grounded in narrow, task-specific environments, and there is a clear need for more general-purpose frameworks and benchmarks that reflect the diversity of real-world collaboration. As AI becomes increasingly integrated into workplace, educational, and social group contexts, developing robust simulation tools for many-human–AI interaction will be essential for building agents that are trustworthy, effective, and contextually aware in group settings.

\section{How: Techniques and Methodologies (Extended)}
\label{sec:appendix-how}
\begin{table*}[ht!]
\centering
\scriptsize
\footnotesize
\caption{Taxonomy of User Simulated Data Generation Techniques.
}
\begin{adjustbox}{width=\textwidth}
\begin{tabular}{@{}llHlH@{}}
\toprule
& 
& \textbf{Definition} & \textbf{Description} & \textbf{Example} \\ 
\midrule
Prompt-based Generation & Zero-shot & $p(y|x) = p_{\theta}(y|x)$ & No examples provided, direct inference & ``Translate `hello' to French.'' \\
& Few-shot & $p(y|x, \{x_i,y_i\}_{i=1}^k) = p_{\theta}(y|x, \{x_i,y_i\}_{i=1}^k)$ & Inference guided by few examples & ``Translate `cat' (example: dog→chien)'' \\
& Chain-of-thought & $p(y, r|x) = p_{\theta}(y|r,x) \cdot p_{\theta}(r|x)$ & Intermediate reasoning steps provided & ``Solve 12×5: First, 10×5=50, then 2×5=10, total is 60.'' \\
\midrule
Fine-tuning Methods & Supervised Fine-tuning (SFT) & $\theta^* = \arg\max_{\theta}\sum_{(x,y)\in D}\log p_{\theta}(y|x)$ & Learning from explicit labeled data & Training sentiment analysis model on labeled reviews \\
& Reinforcement Learning (RLHF) & $\theta^* = \arg\max_{\theta}\mathbb{E}_{x\sim D}[\mathbb{E}_{y\sim p_{\theta}(y|x)}[r_{\phi}(y,x)]]$ & Learning via feedback-driven reward signals & Model learning summarization preferences from human ratings \\
\midrule
Hybrid Approaches & Retrieval-augmented Generation & $p(y|x) = p_{\theta}(y|x,\mathcal{R}(x))$ & Generation enhanced by external information retrieval & Answering questions by retrieving relevant Wikipedia paragraphs \\
& Persona-based Generation & $p(y|x,u) = p_{\theta}(y|x,u)$ & Generation personalized based on user or persona context & Conversational agent responding as a specific character or user persona \\ 
& Persona-based RAG & $p(y|x, u) = p_{\theta}(y|x,\mathcal{R}(x, u), u)$ & Generate based on persona context and RAG  \\
\bottomrule
\end{tabular}
\end{adjustbox}
\end{table*}

\subsection{Prompt‑Based Simulation}\label{append:how-prompt}

\paragraph{Zero‑/Few‑Shot Prompting.}
A zero‑shot prompt provides only high‑level instructions; a few‑shot prompt additionally lists a handful exemplars. \citet{terragni2023context} use few‑shot in‑context examples to
prompt GPT‑NeoX for task‑oriented dialogue (TOD), achieving diverse
responses without fine‑tuning.  
\citet{zhang2024generative}
represent each synthetic recommender‑system user as an LLM endowed with
a profile, memory, and action space, enabling near‑zero‑shot behavioral
simulation.  
\citet{wang2025user} extend this idea to Web environments, showing that
LLM agents can simulate browsing and clicking behaviors with minimal
user data.

\paragraph{Chain‑of‑Thought Prompting.}
CoT prompts ask the model to reason step‑by‑step before producing the
final utterance, thereby improving logical consistency.  
DuetSim \cite{luo2024duetsim} adopts a generator–verifier loop:
one LLM enumerates dialogue acts in CoT form, while a second verifies
and verbalizes them, leading to stronger goal coherence.

\paragraph{Persona / Role‑Play Prompting.}
This prompting strategy assigns the LLM a persona (e.g.\ ``novice buyer''), thereby
controlling tone and constraints \cite{shanahan2023role}.
\citet{abbasiantaeb2024let} run two GPT‑4 agents (student/teacher) to
generate multi‑turn QA dialogues automatically.
\citet{kong2023platolm} propose a Socratic user simulator that generates
more natural, diverse questions than earlier role‑play pipelines such as
Baize~\cite{xu2023baize} and UltraChat~\cite{ding2023enhancing}.

\paragraph{Task‑Specific Prompting.}
These prompts encode detailed domain constraints or required actions so
that generated utterances align tightly with an application.  
\citet{terragni2023context} supply GPT‑4 with a high‑level TOD goal and
a few exemplar dialogues, letting the model adapt across domains.
For conversational search, \citet{kiesel2024simulating} prompt an LLM to ask plausible follow‑up questions given a system answer, producing
queries that match human behavior on both automatic metrics and human
evaluation.

\subsection{RAG} \label{append:how-rag}
We categorize RAG-based simulation techniques by their retrieval trigger mechanism: \textit{Always-on}, \textit{Adaptive}, and \textit{Goal / State–driven}.

\paragraph{Always-on.} Retrieval is applied at every turn, regardless of necessity. For example, KAUCUS\cite{shimadzu2025retrieval} introduces a framework for building diverse user simulators by leveraging external text. Its SRAG (Simulator with RAG) model prepends a retrieved passage to each user prompt. Specifically, for every user turn in the training data, SRAG queries MS MARCO using the previous utterance (via BM25) and prepends the top-ranked passage before generating the next response. This retrieval step enriches the simulator with factual, varied content, enhancing the diversity and realism of simulated behaviors.

\citet{shimadzu2025retrieval} extend this idea to large-scale social network simulations. Each LLM-based agent mimics human behavior by retrieving up-to-date content (e.g., from the web) before generating posts or replies. This retrieval step enables agents to discuss trending topics beyond the LLM's pretraining, leading to more natural and timely interactions. Compared to static simulators, these retrieval-augmented agents better capture dynamic social media behavior.

\paragraph{Adaptive.} Adaptive-gate methods use a learned classifier to decide, at each turn, whether retrieval is necessary. This selective retrieval strategy improves efficiency by avoiding unnecessary queries and reduces noise. For instance, RAGate \cite{wang2024adaptive} introduces an LLM-based gating mechanism trained on human-labeled dialogues to predict whether to retrieve external knowledge. Given the dialogue history and user/goal context, the gate outputs a binary decision—retrieve or not. By augmenting only relevant turns, RAGate improves simulation efficiency.

\paragraph{Goal / State Driven.} Retrieval is guided by the simulator's internal state—such as user goals or memory contents. \citet{zhu2025llm} propose CSHI, a modular simulator for conversational recommendation systems that maintains structured user preference memory. At each turn, CSHI retrieves relevant preferences to generate contextually appropriate responses. For instance, when asked about likes or recommendations, the simulator queries its long-term and real-time memory for matching items or traits. It also dynamically updates its memory as new preferences emerge during the interaction. This goal- and memory-driven retrieval helps ensure coherence and personalization in simulated user behavior.

\subsection{Fine-tuning} \label{append:how-fine-tuning}
We group existing methods into three broad categories: \textit{Full-model supervised}, \textit{Parameter-efficient}, and \textit{Interactive / self-play} fine-tuning.

\paragraph{Full-model supervised.}
These works update \emph{all} weights on in-domain dialogues.
DAUS~\cite{sekulic2024reliable} fine-tunes LLaMA-7B on two task-oriented dialogue (TOD) datasets, halving hallucination rates and improving goal fulfillment.
In the affective domain, SoulChat~\cite{chen2023soulchat} performs full SFT on a 2 M-turn empathy corpus to raise human-rated listening and comfort scores.
MuPaS~\cite{wang2024multi} extends the idea to \emph{multi-party} settings by masking inactive speakers during SFT so a single model can generate coherent group conversations.

\paragraph{Parameter-efficient.}
Instead of updating the full model, these methods insert lightweight adapters.
ESC-Role~\cite{zhao2024esc} LoRA-tunes Qwen-14B on merged ESCConv/ExTES/Smile data to produce realistic emotional-support seekers.
BiPO steering vectors~\cite{cao2024personalized} learn small activation-space adapters that steer an LLM toward target dialogue strategies without touching the backbone.
Another paper~\cite{madani2024steering} combines a Strategy Relevance Assessment (SRA) gate with LoRA fine-tuning: strategy-bearing tokens are detected, and ground-truth strategy-conditioned responses are injected mid-sequence to prevent intention drift in lengthy emotional-support conversations.

\paragraph{Interactive / self-play.}
Here, the simulator is refined with \emph{interaction signals} rather than static targets.
UGRO~\cite{hu2023unlocking} lets an LLM act as a user-side judge that scores candidate replies; those rewards are used to fine-tune the TOD model with PPO, boosting task success.
PlatoLM~\cite{kong2023platolm} first SFTs a 'Socratic' questioner, dialogues it with ChatGPT to create 600 k self-play conversations, then fine-tunes the assistant, achieving MT-Bench SOTA for 7 B models.

\subsection{RL/DPO} \label{append:how-rl-dpo}
We categorize RL/DPO-based simulation techniques by the dimension of conversational adaptation they target: \textit{Personalization}, \textit{Memory}, \textit{Long-Horizon Planning}, \textit{Offline/Hindsight Learning}, and \textit{Action-level Clarification}.

\paragraph{Personalization.}
Personalization-driven RL/DPO encourages simulators to infer and adapt to individual user traits or latent preferences through the course of interaction. For example, \citet{wan2025enhancing} propose a curiosity-based reward in RLHF that motivates the simulator to elicit and disambiguate latent user types. The policy is rewarded for actions that sharpen its belief over the user's profile, resulting in more tailored and proactive questioning during multi-turn dialogues. This approach outperforms standard RLHF on simulated user personalization tasks.

\paragraph{Memory.}
Memory-aware RL/DPO methods optimize what content from past dialogue should be retained or recalled to maximize future conversational quality.
\citet{seo2024efficient} introduce a DPO-based memory selector that dynamically updates the simulator's external memory. By learning from preferences over different memory configurations, the simulator achieves improved factual recall and coherence in follow-up turns, while maintaining a compact memory state.

\paragraph{Long-Horizon Planning.}
Long-horizon RL approaches address the challenge of assigning credit and planning over many dialogue turns. \citet{zhou2024archer} present ArCHer, a hierarchical RL method with a high-level Q-learning policy for utterance-level goals and a low-level PPO agent for token-level realization. This structure enables the simulator to optimize behaviors spanning dozens of turns, drastically improving sample efficiency and strategic coherence in multi-turn scenarios such as tool use or customer support.

\paragraph{Offline and Hindsight Learning.}
Offline and hindsight-based RL methods improve simulator policy using retrospective credit assignment or offline data augmentation, reducing the need for costly interactive roll-outs.
\citet{hong2024interactive} employ ``hindsight regeneration'' to rewrite suboptimal dialogue segments after observing user reactions, then fine-tune the simulator offline on these augmented interactions. This yields more effective user steering in counseling and persuasion settings compared to standard imitation or prompt-based approaches.

\paragraph{Action-level Clarification.}
Action-level DPO techniques optimize the simulator's ability to resolve ambiguity by learning when and how to clarify user intent.
\citet{chen2024learning} frame clarification as a preference optimization problem—contrasting possible next actions (clarifying vs. answering vs. delegating)—and train the simulator to prefer actions that maximize downstream task success and user satisfaction. Their method achieves higher performance on multi-turn QA and goal-oriented dialogue tasks relative to standard supervised or DPO training.

\subsection{Hybrid Approaches} \label{append:how-hybrid}
We organize hybrid simulation techniques by the integration pattern they employ:

\paragraph{Retrieval-Augmented Fine-Tuning.}
Some approaches inject retrieved passages \textit{during fine-tuning} instead of only at inference. For example, ~\citet{zhang2024raft} introduce RAFT, which fine-tunes LLaMA‑derived models using triplets of query, retrieved passages (including distractors) and chain-of-thought answers, training the model to cite relevant evidence and disregard irrelevant context while improving performance on domain-specific benchmarks such as PubMed, HotpotQA, and Gorilla \cite{zhang2024raft}. Similarly, ~\citet{kaiser2025preference} propose PRAISE for conversational QA: it trains modular adapters through Direct Preference Optimization by iteratively generating, ranking, and fine-tuning on top-ranked (positive) versus lower-ranked (negative) context–answer samples, enhancing factual grounding and response robustness \cite{kaiser2025preference}. These hybrid retrieval-and-fine-tuning methods outperform both standard RAG and vanilla supervised fine-tuning in terms of factual accuracy and conversational resilience.

\paragraph{Prompt-to-Fine-Tune Curricula.}
A common hybrid strategy uses prompt-based simulators to bootstrap high-quality synthetic data, then fine-tunes adapters or the full model on this expanded corpus. \citet{chen2023soulchat} employ zero-shot GPT-4 role-play to generate millions of empathetic dialogues, followed by LoRA-based fine-tuning for fast, domain-adapted simulation. \citet{kong2023platolm} use Socratic self-play to create hundreds of thousands of multi-turn conversations, then apply supervised fine-tuning to stabilize persona and dialogue quality. This curriculum leverages the rapid adaptability of prompts and the efficiency and consistency of finetuned models.

\paragraph{RAG + RL/DPO Loops.}
Here, retrieval mechanisms are coupled with RL/DPO policies: the RL/DPO policy decides whether to retrieve, how to incorporate the retrieved context, and how to adjust simulator strategy based on feedback. MemDPO \cite{seo2024efficient} learns a memory selector using preference optimization, enabling the simulator to maintain a compact, relevant external memory for better recall and coherence. Curiosity-RAGate combines curiosity-driven RLHF rewards (see §\ref{sec:how-rl-dpo}) with adaptive retrieval, encouraging clarification queries only when the user's preferences are ambiguous \cite{wan2025enhancing}.

\paragraph{Hierarchical Modular Pipelines.}
Some hybrid systems structure the simulation pipeline as a set of specialized modules, each using different methods. For example, ARCHer \cite{zhou2024archer} (see §\ref{sec:how-rl-dpo}) employs a planner module with CoT prompting, a retriever module for web or tool queries, a finetuned executor module, and a DPO-trained critic, each collaborating to optimize long-horizon dialogue. Social-network simulators like KAUCUS-SRAG integrate always-on retrieval and RL-tuned posting policies to simulate dynamic trends \cite{shimadzu2025retrieval}.

\paragraph{Personalized Hybrid Stacks.}
Cutting-edge personalized simulators operate on multiple levels: using persona-driven prompting, private memory retrieval, adapter-based fine-tuning for user-specific style, and RLHF for preference alignment. Recent surveys (see, e.g., \citet{zhang2024personalization, liu2025survey}) describe this as a three-tier system—input (prompt/RAG), model (parameter-efficient fine-tuning), and objective (RL/DPO alignment)—that has become standard in open-source frameworks.

\section{Evaluation (Extended)}
\subsection{Traditional Metrics}
\label{app:traditional-metrics}
\paragraph{N-gram overlap.} 
BLEU/ROUGE-style metrics are still reported in several dialog simulation settings (e.g., BLEU/F1 in SimDial \cite{zhao2018zero}, ROUGE/F1 in DialogStudio \cite{zhang2023dialogstudio}, ROUGE-L in RoleLLM \cite{wang2023rolellm}). 
They are easy to compute but correlate poorly with human judgments in open-ended conversation, where many acceptable responses share little lexical overlap with references.

\paragraph{F1 slots / fact accuracy.} 
When a dialogue contains well-defined, enumerable facts or slots (e.g., Wizard-of-Oz style tasks such as CrossWOZ \cite{zhu2020crosswoz} or MultiWOZ \cite{ye2021multiwoz}), precision/recall-style measures (F1) or exact match on slots/entities provide a clear signal for information correctness, but they ignore pragmatic qualities such as empathy or persona fidelity.

\paragraph{Perplexity.} Perplexity~\cite{chen1998evaluation} is sometimes reported to indicate distributional fit to a human corpus, but it is an intrinsic language model measure and does not necessarily reflect interactive quality or human preference.

\paragraph{Task success / accuracy.} In goal-driven simulations such as negotiation \cite{lewis2017deal,he2018decoupling} or action decision \cite{chen2021action}, binary or graded success criteria are natural. 
These metrics are interpretable but only apply when a clear ground-truth objective is defined.

In practice, these traditional metrics offer low-cost, reproducible signals, but they measure narrow facets of dialogue quality. 
As a result, many recent works pair them with LLM or human judges to capture semantic and pragmatic dimensions that overlap metrics miss.
\subsection{Human Evaluation}
\label{append:human-evaluation}
Human evaluation remains the reference standard. 
Two modes are prevalent. 
The first is \emph{interactive} evaluation, where humans converse with a system (possibly paired with a simulator) and then rate satisfaction, coherence, or realism. 
The second is \emph{offline} evaluation, where annotators read transcripts and provide Likert-scale ratings or pairwise preferences over full conversations or single turns. 
For example, MultiWOZ \cite{ye2021multiwoz} reports human judgments on dialogue quality; role/character benchmarks such as CharacterBench \cite{zhou2025characterbench} and LifeStageBench \cite{fan2025if} also rely on expert or crowd annotators for final scoring, sometimes in combination with model judges.
Common protocols include: (i) Likert scoring on multiple axes (e.g., naturalness, coherence, goal completion, persona/role fidelity), (ii) pairwise A/B testing that asks which conversation (or response) is better along one criterion, and (iii) reporting inter-annotator agreement to quantify rating consistency. 

The main downsides of human evaluation are cost, latency, and limited reproducibility across studies due to differences in annotator pools, rubrics, and scales.
These issues motivate the growing use of LLM-as-Judge as a scalable proxy, followed by \emph{meta-evaluation} that quantifies how well model-judged scores align with human ratings on shared subsets \cite{fan2025if,zhou2025characterbench}. 
Nevertheless, for high-stakes or nuanced aspects such as safety, subtle persona drift, or social norm adherence, human studies remain the final arbiter.

\section{Datasets (Extended)}
\label{append:datasets}
\paragraph{Personalized conversations.}
Personalized datasets~\cite{li2025personalized,zheng2019personalized,zhang2018personalizing} focuses on natural conversations in various domains.
These conversations happen between individuals of diverse traits, including age, gender, location, and personal interests, among others.

\paragraph{Multiparty dialogues.}
Multiparty dialogues~\cite{gao2023livechat,liu2023one} are common on social platforms, where usually one person serves as a host or lead talker, while others engage in the conversations.
These datasets usually involve addressee recognition to determine whom the host responds to.

\paragraph{Question answering.}
Question answering datasets contain conversations between two persons, one asking questions, the other answering.
SimQuAC~\cite{abbasiantaeb2024let} consists of LLM-based simulated teacher-student conversations.
These conversations focus on question-answering over Wikipedia articles, where the student asks questions to explore the article's topic, and the teacher responds based on the article.
SocraticChat~\cite{kong2023platolm} contain conversations of simulated human-like questions.
Furthermore, several datasets~\cite{zhou2025characterbench,shao2023character,tu2024charactereval,wang2023rolellm} contain dialogues where simulated responses are given based on characters from novels or TV shows.

\paragraph{Ranking.}
Nectar~\cite{zhu2024starling} contains conversations from various LLMs.
The dataset is used for response ranking.

\paragraph{Multi-domain dialogues.}
SimDial~\cite{zhao2018zero} contains dialogues in multiple scenarios, including restaurants, movie, bus, and weather, among others.

\paragraph{Wizard-of-Oz.}
Wizard-of-Oz datasets~\cite{byrne2019taskmaster,zhu2020crosswoz,ye2021multiwoz} contains conversations between humans to help develop future conversational systems, or wizards.
Each conversation contains two role, users and wizards, where users ask questions for a given goal, wizards give answers according to the questions.

\paragraph{Memory enhancement.}
A few datasets~\cite{fan2025if,chen2024socialbench} consists of dialogues that are generated by LLMs with focus on long-term conversation memory.
Hence, these conversations are usually longer and have more connections among different turns.

\paragraph{Negotiation.}
Negotiation datasets~\cite{he2018decoupling,lewis2017deal} consist of dialogues between two persons, where in DealOrNoDeal~\cite{lewis2017deal} they are asked to divide a set of items, and in CraigslistBargain one person serves as a buyer to negotiate down item prices, while the other serves as a seller to profit as much as possible.
\begin{table*}[ht!]
\centering
\scriptsize
\caption{%
Summary of Datasets.
Taxonomy of User Simulated Conversational \& Other Complex Datasets.
Note that 
what
($\S$\ref{sec:what}) is 
Human-AI $\S$\ref{sec:what-human-ai} (H-AI), 
AI-AI ($\S$\ref{sec:what-ai-ai}), 
Human-Human $\S$\ref{sec:what-human-human} (H-H), and so on.
}
\rowcolors{2}{googlegray!20}{white}
\label{table:datasets}
\begin{adjustbox}{width=\textwidth}
\begin{tabular}{lccHllHcHcc}
\toprule
& \textbf{Who} & 
\textbf{What} & 
\textbf{Data Type} & \textbf{Task} & 
\textbf{Data Size} & \textbf{Test} & 
\textbf{Eval. Metric} & \textbf{Approach} & \textbf{Code} \\ 
\midrule

PersonaChat\cite{zhang2018personalizing} & Indiv. & H-H & List & Text Gen. & 11K Conv. & 1K Conv. & \makecell[c]{Likelihood / F1 / \\ Classification Loss} & -- & \href{https://huggingface.co/datasets/awsaf49/persona-chat}{[code]}\\

PersonalDialog~\cite{zheng2019personalized} & Indiv. & H-H & List & Text Gen. & 21M Conv. & - & Perplexity / Accuracy & -- & \href{mailto:zhengyinhe1@163.com}{[link]}\\

PersonalConv~\cite{li2025personalized} & Indiv. & H-H & List & \makecell[l]{Classif., \\ Regres., Gen.} & 111K Conv. & - & \makecell[c]{Accuracy / F1 / MCC / \\ MAE / RMSE / \\ ROUGE-1 / ROUGE-L / \\ BLEU / METEOR}  & -- & \href{https://github.com/PERSONA-bench/PERSONA/tree/Latest}{[code]}\\

\midrule
LiveChat~\cite{gao2023livechat} & Persona & H-H & List & Gen., Recog. & 1M Conv. & - & \makecell[c]{Recall / MRR / BLEU-n / \\ ROUGE-n / ROUGE-L} & - & \href{https://github.com/gaojingsheng/LiveChat}{[code]}\\

\midrule

SimDial~\cite{zhao2018zero} & Gen. & AI-AI & List & Dialog Gen. & 9K Conv. & - & BLEU / F1 & -- & \href{https://github.com/snakeztc/SimDial}{[code]}\\

Taskmaster-1~\cite{byrne2019taskmaster} & Gen. & H-H / H-AI & List & Wizard-of-Oz & 13K Conv. & - & Human Judge & -- & \href{https://github.com/google-research-datasets/Taskmaster}{[code]}\\

CrossWOZ~\cite{zhu2020crosswoz} & Gen. & H-H & List & Wizard-of-Oz & 6K Conv. & - & F1 & -- & \href{https://github.com/thu-coai/CrossWOZ}{[code]}\\

MultiWOZ~\cite{ye2021multiwoz} & Gen. & H-H & List & Wizard-of-Oz & 10K Conv. & - & Human Judge & -- & \href{https://github.com/smartyfh/MultiWOZ2.4}{[code]}\\

SimQuAC~\cite{abbasiantaeb2024let} & Gen. & AI-AI & - & Q\&A & \makecell[c]{334 Conv. \\ 4K Ques.} & - & Human Judge & Zero-shot student-teacher model & \href{https://github.com/ZahraAbbasiantaeb/SimQUAC}{[code]} \\

SocraticChat~\cite{kong2023platolm} & Gen. & H-AI & List & Q\&A & 25K Conv. & - & LLM Judge & - & \href{https://github.com/FreedomIntelligence/PlatoLM}{[code]}\\

Nectar~\cite{zhu2024starling} & Gen. & AI-AI & List & Conversations & 183K Prompts& - & - & Ranking & \href{https://github.com/efrick2002/Starling}{[code]}\\

\midrule

DealOrNoDeal~\cite{lewis2017deal} & Role & H-H & List & Negotiation & 6K Conv. & - & Human Judge & -- & \href{https://github.com/facebookresearch/end-to-end-negotiator}{[code]}\\

CraigslistBargain~\cite{he2018decoupling} & Role & H-H & List & Negotiation & 7K Conv. & - & Human Judge & -- & \href{https://github.com/stanfordnlp/cocoa/tree/master/craigslistbargain}{[code]}\\

ABCD~\cite{chen2021action} & Role & H-H & List & Action Decision & 10 K Conv. & - & Recall & -- & \href{https://github.com/asappresearch/abcd}{[code]}\\

SAD~\cite{liu2023one} & Role & AI-AI & List & Q\&A & 6K & - & \makecell[c]{LLM / \\ Automatic Evaluation} & -- & \href{https://github.com/kiseliu/must}{[code]}\\

RoleLLM~\cite{wang2023rolellm} & Role & H-AI & List & Q\&A & 168K S. & - & ROUGE-L & -- & \href{https://github.com/InteractiveNLP-Team/RoleLLM-public}{[code]}\\

CharacterLLM~\cite{shao2023character} & Role & H-H / H-AI & List & \makecell[l]{Gen.,\\ Interview Q\&A} & 14.4K scenes & - & LLM Judge & -- & \href{https://github.com/choosewhatulike/trainable-agents}{[code]}\\

SocialBench~\cite{chen2024socialbench} & Role & ALL & List & Q\&A & 6K Ques. & - & Acc / Cover & -- & \href{https://github.com/X-PLUG/SocialBench}{[code]}\\

CharacterEval~\cite{tu2024charactereval} & Role & H-H & List & Q\&A & 2K Conv. & - & Custom & -- & \href{https://github.com/morecry/CharacterEval}{[code]}\\

CharacterBench~\cite{zhou2025characterbench} & Role & H-H & List & Q\&A & 23K S. & - & Model Judge & -- & \href{https://github.com/thu-coai/CharacterBench}{[code]}\\

LifeStageBench~\cite{fan2025if} & Role & AI-AI & List & Role-based Q\&A & 1.3K / 202 S. & - & Model Judge & -- & -- \\

\midrule

DialogStudio~\cite{zhang2023dialogstudio} & Hybrid & ALL & List & - & Union & - & ROUGE-L / F1 & -- & \href{https://github.com/salesforce/DialogStudio}{[code]}\\

\bottomrule
\end{tabular}
\end{adjustbox}
\end{table*}

\section{Applications (Extended)}
\label{app:application}
\paragraph{Scalable Evaluation of Dialogue Agents.}
Traditional user studies are expensive and difficult to replicate. Simulated users offer an efficient and reproducible alternative for evaluating dialogue systems across diverse scenarios, including rare edge cases. By configuring user profiles and behavioral trajectories, researchers can systematically assess model performance under controlled yet realistic conditions. This is particularly useful in domains such as personalized customer support \cite{patel2020leveraging}, where user-specific behavior must be consistently modeled across sessions.

\paragraph{Training Robust and Adaptive Systems.}
In reinforcement learning or imitation learning settings, user simulators can generate large-scale, contextually rich interactions without requiring live deployment. This facilitates closed-loop training where conversational agents learn to handle ambiguity, recover from errors, and generalize across user intents. Simulation also supports curriculum learning, where complexity can be increased progressively. Applications include adaptive virtual assistants that personalize their behavior over time based on user history \cite{lamontagne2014framework}.

\paragraph{Design Exploration and Prototyping.}
Before real users are involved, system designers often need to explore how different dialogue policies behave under varied user goals or engagement patterns. Simulators can be configured to emulate distinct user types, such as cooperative users, those who frequently disengage, or users who ask clarifying or inquisitive questions. This supports early-stage testing of system responses, decision timing, and personalization strategies. In educational technology, for example, simulated students with varying misconceptions or engagement levels can inform the development of personalized tutoring systems \cite{lin2023artificial}.

\paragraph{Research on Human-Like Behavior and Alignment.}
User simulation is not only a tool for engineering systems but also a methodological asset for studying human-machine interaction. Simulated dialogues can serve as proxies for human behavioral data, enabling experiments on alignment, fairness, or affective response modeling. For instance, emotional dynamics and social support behaviors can be explored through simulated users in wellness-oriented dialogue settings \cite{tutun2023ai, li2024panoptic}.

\paragraph{Benchmark Construction and Diagnostic Analysis.}
Finally, LLM-driven simulators can be used to automatically generate structured benchmarks that test specific capabilities of conversational agents, such as contextual recall, politeness strategies, or belief tracking. Simulation also aids in generating counterfactual examples and behavioral probes, which are valuable for identifying model blind spots and diagnosing failure modes.

\subsection{Recommendation}
Conversational user simulation has become a pivotal tool in developing, training, and evaluating recommendation systems, enabling scalable and controlled experimentation without requiring extensive human interaction. A foundational contribution by \cite{zhang2020evaluating} introduced an agenda-based user simulator tailored for evaluating dialogue flows in recommendation systems, demonstrating that simulated evaluations could closely mirror human judgment. Building on this, ~\cite{afzali2023usersimcrs} developed UserSimCRS, an extensible simulation toolkit that incorporates user personas, satisfaction prediction, and conditional language generation, further enhancing realism and adaptability in recommendation contexts such as movie dialogues. To move beyond general evaluation and into system diagnostics, ~\citet{bernard2024identifying} proposed a simulator-driven framework to identify conversational breakdowns in conversational recommendation system interactions, offering a novel approach to robustness testing. With the advent of large language models (LLMs), more recent work has explored generative user simulators. ~\citet{zhu2025llm} presented a controllable, scalable LLM-based simulator framework (CSHI) that supports plugin-based behavior modulation and personalization, while ~\cite{yoon2024evaluating} introduced a five-task evaluation protocol to systematically benchmark LLMs' ability to simulate user behavior in recommendation dialogues, revealing both potential and limitations (e.g., popularity bias, weak personalization). Together, these studies demonstrate that user simulation not only supports evaluation but also actively contributes to the design, optimization, and understanding of conversational recommendation systems.

\subsection{Education} 
With the rise and evolution of LLMs, the fusion of AI and education has entered a new phase~\cite{wang2024large, xu2024large}, particularly in the domain of conversational user simulation. The current research shows that the simulation applications in education are widely adopting Generative AI (GenAI), with a significant reliance on closed-source LLMs represented by OpenAI's GPT family~\cite{achiam2023gpt}. The public release of ChatGPT in late 2022 greatly catalyzed this trend; the advanced conversational generation abilities of such LLMs are valued for ease of integration and high performance, and have been rapidly applied in various educational scenarios~\cite{park2024ai, zhang2024simulating}. However, there's also exploration into open-sourced alternatives~\cite{xu2024large, hohn2024beyond, lee2024developing}. This shift is driven by the pursuit of greater customization~\cite{jin2025teachtune}, data security, and transparency—features that allow models to be more deeply adapted to specific pedagogical contexts~\cite{zheng2025teaching}. Such adaptation efforts often go beyond simple prompt engineering, instead involving fine-tuning on domain-specific data to align the AI's simulated user behavior with precise teaching objectives.

Furthermore, the role AI plays in educational user simulations has undergone a fundamental transformation. In the pre-GenAI era, simulated user behavior was often limited and predefined by pre-set scripts. LLMs enable AI-powered simulator to embody a much broader and more dynamic range of user characteristics, thereby creating more interactive and adaptive learning experiences. An intuitive thinking would be simulating a teacher, a tutor, a teaching assistant or coach~\cite{olkedzka2024ai, ye2025position}, especially for language teaching or social skill training~\cite{park2024ai, lee2024developing}. However, when it comes to math or physics, for instance, a more practical application is the use of LLMs to simulate students~\cite{yue2024mathvc, pan2025tutorup} due to their limited capabilities~\cite{frieder2023mathematical}. These simulations provide a platform for pre-service teachers to practice classroom management and instructional strategies in a low-stakes environment~\cite{pan2025tutorup, lim2025development}. A key research frontier in this area is enhancing the realism of these simulated students, by modeling the behavior of students at different cognitive levels, including common misconceptions and error patterns, to create more authentic and challenging training scenarios~\cite{yue2024mathvc}. Beyond simulating students, LLMs are also being used to simulate learning partners or collaborators to support tasks such as collaborative problem-solving or language practice~\cite{lytvyn2025human}.

\subsection{Limited User Feedback}
In real conversations, user feedback is sparse, delayed, or ambiguous, making it hard to train and evaluate user simulators based on clean supervision. Most human-annotated datasets rely on intensive labeling efforts \cite{jang2022call,ahn2023mpchat} and are limited in size and diversity \cite{wang2022self,xu2024wizardlm}. As a result, existing works often rely on data that implicitly reflects idealized user behavior, providing dense, turn-level signals and overlooking phenomena like dropout, disengagement, or subtle feedback cues. Addressing this challenge may require integrating synthetic and human-authored data, modeling implicit or delayed feedback, and exploring learning objectives that better tolerate sparse and noisy supervision.

\subsection{HCI/UI}
User-conversational simulated data has significant implications for the fields of human-computer interaction and interface design \cite{moore2018conversational}. On paper, user testing is a crucial step in the user interface design process, but in practice, resources are often so constrained that testing each interface is impossible. Such simulated data offers the opportunity for user researchers to conduct full-scale usability studies \cite{baxter2015understanding, steen2007early}, user surveys, and focus groups without using the extensive resources it currently takes to do so. Furthermore, many corporate user research teams are far outnumbered by designers, with many teams having a ratio of 1:5 \citep{kaplan2020typical}. Given the resource constraints, utilizing user-simulated data in UI testing would allow designers to take part in an integral step that they are often forced to skip. 

Furthermore, studies like \cite{hamalainen2023evaluating} evaluate LLMs' ability to generate synthetic user research data for usability tasks. They underscore the importance of careful prompt writing and using these methods earlier in the design process for need finding and early feedback. The same study even found that LLM simulated conversational data was often distinguishable from human results, with humans actually believing LLM answers to be more ``human'' than the actual human data it was being compared to. Overall, user-conversational simulated data offers the opportunity to conduct HCI and user research in scenarios where research would have initially been too difficult to conduct.

\subsection{Video Understanding} 
Conversational user simulation can also be applied to video understanding domain. VideoAutoArena \cite{luo2025videoautoarena} introduces a novel user of conversational user simulation by generating open-ended adaptive questions to evaluate large multimodal models. Instead of relying on static multiple-choice benchmarks, this framework simulates realistic user inquiries about video content and challenges models through an arena-style evaluation with a modified ELO Rating System. This approach demonstrates that LLM‑based conversational simulation can serve not only as a generator of user-style queries, but also as a rigorous evaluation mechanism, enabling scalable and user-centered assessment of video comprehension in multimodal systems.

\subsection{Discussion}
Admittedly, the scope of conversational user simulation spans a wide array of domains, and it's unrealistic to cover them all. We instead focus on areas where conversation simulation has shown tangible, direct high-impact: (i) augmenting training and evaluation pipelines, (ii) enabling controlled experimentation at scale, and (iii) facilitating personalized interaction design. These domains not only benefit from simulation's scalability but also expose its role in advancing robust, adaptive, and user-centered systems.

\section{Challenges (Extended)}\label{app:challenges}

\subsection{Evaluation}
In the evaluation of conversational user simulation, human judgment remains the gold standard for assessing qualities like coherence, persona fidelity, and goal alignment. However, it is costly, slow, and often inconsistent across annotator pools and rubrics. LLMs are increasingly used as automated judges, but they are often sensitive to prompt wording and lack grounding, calibration, and reliability, particularly for long or complex conversations. Recent work \cite{fan2025if,zhou2025characterbench} has introduced promising practices such as symmetric prompting, ensembling, and meta-evaluation against human ratings, but evaluation protocols still vary across studies. Building standardized, multi-layered evaluation pipelines remains an important open direction.
\subsection{User-Specific Personalization}
Simulating specific users enables personalized training and evaluation, but introduces unique challenges. First, user-level simulation raises privacy concerns, especially when behavioral traces are derived from real individuals. Future work may explore privacy-preserving techniques such as differential privacy, federated learning, or privacy-aware user simulation. Second, real users evolve over time, shifting goals, language use, or engagement patterns, yet most simulators remain static. Third, interaction histories are often limited, making it difficult to capture fine-grained behavioral patterns. Finally, existing benchmarks rarely support user-level evaluation or isolate individual variation in a consistent way. Addressing these challenges calls for methods that can generalize from limited signals, model user drift, and preserve privacy while supporting personalization.

\subsection{Video Understanding}
While recent frameworks like VideoAutoArena \cite{luo2025videoautoarena} showcase the promise of LLM-based conversational simulators for video understanding, several limitations remain. First, aligning simulated user queries with the fine-grained temporal and semantic aspects of videos remains difficult, often leading to generic or misaligned questions. Additionally, maintaining coherent multi-turn conversations about dynamic visual content challenges current LLMs, which struggle with long-range context tracking. Another critical gap lies in the lack of alignment with real-world user behaviors and preferences, limiting the realism and utility of the simulated interactions. Addressing these challenges is essential for deploying truly user-centric conversational agents in video-rich environments.

\paragraph{Evaluation of Dialogue Diversity.}
Quantifying diversity in simulated conversations remains an open challenge. Existing evaluation metrics primarily focus on surface-level textual variation or semantic similarity, which fail to capture deeper diversity in knowledge, reasoning patterns, and persona-specific perspectives. Developing metrics that can evaluate \emph{knowledge diversity} and behavioral variation across simulated characters is essential for assessing the realism and utility of conversational simulations.

\paragraph{Persona Consistency and Adaptability.}
Maintaining persona consistency over long conversations while allowing adaptive responses to evolving contexts presents a fundamental trade-off. Overemphasizing consistency may result in rigid and repetitive behavior, whereas excessive adaptability can lead to persona drift. How to dynamically balance these competing objectives—particularly in multi-turn or open-ended interactions—remains largely unexplored.

\paragraph{Hybrid Simulation Paradigms.}
Existing conversational simulation paradigms span human--AI, AI--AI, and mixed hybrid settings. While each paradigm offers unique advantages, systematic comparisons across these settings are scarce. The lack of unified benchmarks and evaluation protocols hinders a comprehensive understanding of when and how different simulation paradigms should be employed.

\paragraph{Knowledge Updating and Temporal Evolution.}
Most current role-play simulations focus on static personas, such as historical figures or fictional characters. In contrast, simulating active real-world figures requires models to adapt as underlying knowledge and public roles evolve over time. Although prior work has explored adaptive and evolving agents~\cite{evolving_agents}, applying such strategies to conversational role-play—while preserving persona coherence—remains an open research direction.

\paragraph{Bias and Safety in Persona Simulation.}
Persona-based simulation introduces inherent risks of stereotyping, bias amplification, and unsafe behaviors. Mitigating these risks without sacrificing authenticity and expressiveness is a critical challenge. Future work must explore principled approaches to bias detection, controllable generation, and safety-aware persona modeling to ensure responsible deployment of conversational simulators.

\end{document}